\documentclass{article}

\usepackage{arxiv}

\usepackage[utf8]{inputenc} % allow utf-8 input
\usepackage[T1]{fontenc}    % use 8-bit T1 fonts
\usepackage{hyperref}       % hyperlinks
\usepackage{url}            % simple URL typesetting
\usepackage{booktabs}       % professional-quality tables
\usepackage{amsfonts}       % blackboard math symbols
\usepackage{nicefrac}       % compact symbols for 1/2, etc.
\usepackage{microtype}      % microtypography
\usepackage{xcolor}         % colors
\usepackage{graphicx}
\usepackage{subfigure}
\usepackage{amsbsy,amsmath,bbm}
\usepackage{multirow}
\usepackage{hyperref}
\usepackage{wrapfig}
\usepackage[ruled,vlined]{algorithm2e}

\input{defs.set}

\title{Hybrid Generative-Contrastive Representation Learning}

\author{
    Saehoon Kim \\
	Kakao Brain\\
	\texttt{shkim@kakaobrain.com} \\
	\And
	Sungwoong Kim \\
    Kakao Brain \\
    \texttt{swkim@kakaobrain.com} \\
	\And
	Juho Lee \\
	KAIST \\
	\texttt{juholee@kaist.ac.kr} \\
}

\date{}

\renewcommand{\headeright}{}
\renewcommand{\undertitle}{}

\hypersetup{
pdftitle={Hybrid Generative-Contrastive Representation Learning},
pdfauthor={Saehoon Kim, Sungwoong Kim, Juho Lee},
pdfkeywords={Hybrid Learning, Generative Pretraining, Contrastive Learning},
}

\begin{document}
\maketitle

\begin{abstract}
Unsupervised representation learning has recently received lots of interest due to its powerful generalizability through effectively leveraging large-scale unlabeled data. There are two prevalent approaches for this, \emph{contrastive learning} and \emph{generative pre-training}, where the former learns representations from instance-wise discrimination tasks and the latter learns them from estimating the likelihood. These seemingly orthogonal approaches have their own strengths and weaknesses. Contrastive learning tends to extract semantic information and discards details irrelevant for classifying objects, making the representations effective for discriminative tasks while degrading robustness to out-of-distribution data. On the other hand, the generative pre-training directly estimates the data distribution, so the representations tend to be robust but not optimal for discriminative tasks.
In this paper, we show that we could achieve the best of both worlds by a hybrid training scheme.
Specifically, we demonstrated that a transformer-based encoder-decoder architecture trained with both contrastive and generative losses can learn highly discriminative and robust representations without hurting the generative performance. We extensively validate our approach on various tasks. 
Code will be available at \url{https://github.com/kakaobrain/gcrl}.
\end{abstract}

% keywords can be removed

\section{Introduction}
\label{sec:intro}

Learning representations without human annotation has recently achieved remarkable progress, especially in natural language processing and computer vision \cite{Devlin2018arxiv, Lan2019arxiv, Liu2019arxiv, Joshi2019arxiv, Kong2019arxiv, Clark2020iclr, Radford2018, Radford2019, Brown2020arxiv, Lewis2019arxiv, Oord2018arxiv, Hjelm2019iclr, Misra2020cvpr, He2020cvpr, Chen2020arxiv, ChenT2020icml, Caron2020neurips, Grill2020neurips, Tian2020arxiv, Chen2020icml, Donahue2019nips}. Many unsupervised representation learning algorithms have been proposed, and these can be broadly categorized into either generative or self-supervised learning algorithms. Generative learning usually aims to obtain representations that can reconstruct an input using an encoder and a decoder while self-supervised learning generally trains an encoder to solve a pretext task derived from unlabeled data\footnote{Generative learning can be considered as a self-supervised learning~\cite{xiao2020arxiv}, however, we separate it that performs an explicit decoding process to the input space.}. In natural language processing, both generative learning and self-supervised learning are widely used for unsupervised representation learning. In computer vision, self-supervised learning, especially contrastive learning~\cite{Oord2018arxiv,ChenT2020icml} based on strong augmentations, is mostly adopted, and generative pre-training~\cite{Chen2020icml, Donahue2019nips} has recently shown some progress.
%However, in computer vision, self-supervised learning especially based on a contrastive objective over strong augmentations, called contrastive learning \cite{Oord2018arxiv, ChenT2020icml}, is mostly adopted even though some recent studies exploit generative pre-training \cite{Chen2020icml, Donahue2019nips}.

For downstream discriminative vision tasks such as image classification, contrastive learning has shown better performances than generative pre-training since it is trained with pretext tasks requiring instance discrimination, and thus encouraged to learn semantic representations rather than minor details. However, since contrastive learning only learns to discriminate images by their identities, it may be less robust under distributional shifts, e.g., less calibrated for out-of-distribution data or low-data transfer settings~\cite{winkens2020contrastive,Ericsson2020arxiv}. On the other hand, the representations learned from generative pre-training may not be as efficient as the ones from contrastive learning, they are more likely to be robust under distributional shift or low-data regime. In addition, generative learning can reduce the reliance on the manually designed pretext tasks or augmentations which often incur overfitting problems \cite{Tamkin2021iclr, Lee2020arxiv}. Similar to the trade-off between the discriminative and generative modeling \cite{Xue2010, Mackowiak2020arxiv, Grathwohl2020iclr}, these two unsupervised representation learning objectives seem to be orthogonal and moreover incompatible, and hence there is almost no existing work to reap the benefits of both objectives in a multi-task pre-training way.

\iffalse
For downstream computer vision tasks that are discriminative such as image classification,
contrastive learning has shown better performances compared to generative pre-training since
its pretext task of instance discrimination leads to produce more discriminative semantic representations without retaining detailed information. However, this discriminative nature of contrastive learning has a limitation in representation capabilities of out-of-distribution detection, uncertainty calibration, and low-data transfer \cite{winkens2020contrastive, Ericsson2020arxiv}. On the other hand, generative learning directly estimates the data distribution, and therefore the obtained representations can be suboptimal for discriminative tasks but more robust to out-of-distribution samples with accurately calibrated uncertainty. In addition, generative learning can reduce the reliance on the manually designed pretext tasks or augmentations which often incur overfitting problems \cite{Tamkin2021iclr, Lee2020arxiv}. Similar to the trade-off between the discriminative and generative modeling \cite{Xue2010, Mackowiak2020arxiv, Grathwohl2020iclr}, these two unsupervised representation learning objectives seem to be orthogonal and moreover incompatible, and hence there is almost no existing work to reap the benefits of both objectives in a multi-task pre-training way.
\fi

In this paper, we propose a hybrid multi-task learning framework that can achieve the merits of both generative and contrastive representation learning. In particular, while maintaining the structure of generative pre-training composed of autoregressive transformer blocks~\cite{Vaswani2017nips, Chen2020icml} since they pose minimal inductive biases and thus are effective for generative modeling, we introduce an encoder-decoder architecture to explicitly separate the role of the blocks. Then, the instance-wise contrastive loss is applied to the pooled representations from the encoder while the generative loss is imposed on the output of the decoder. This separation alleviates the trade-off between the two objectives, and thus enables the encoder to learn both discriminative and robust representations. Experimental results on various image classification benchmarks show that the proposed hybrid approach, which we call Generative-Contrastive Representation Learning (GCRL), outperforms both the generative pre-training and contrastive learning when applied to downstream classification tasks with linear evaluation as well as out-of-distribution detection tasks. In addition, GCRL improves calibration of the prediction uncertainty and performance on low-shot transfer tasks. Furthermore, GCRL does not decrease generative performances of the decoder. Our main contributions can be summarized as follows:

\iffalse
In this paper, we propose to achieve the merits of both generative and contrastive representation learning by introducing a hybrid multi-task learning framework. In particular, instead of solely utilizing the encoder network we construct the encoder and the decoder networks where both are composed of the same autoregressive transformer blocks \cite{Vaswani2017nips, Chen2020icml}. This transformer network relieves the inductive bias on the architecture and enables enhanced generative modeling. Then, the instance-wise contrastive loss is applied to the pooled representation of the encoder while the generative loss is imposed on the image prediction of the decoder. Rather than simultaneously applying the two losses to the same layer, this layer separation for each loss alleviates the trade-off between the two objectives, and therefore, the pre-trained representation from our encoder is able to be not only highly discriminative but also robust to out-of-distribution samples. Experimental results on various image classification benchmarks show that the proposed hybrid generative-contrastive representation learning, which we call GCRL, outperforms both the generative pre-training and the contrastive learning when conducting downstream linear evaluation as well as out-of-distribution detection. In addition, GCRL improves calibration of the prediction uncertainty and performances on low-shot transfer tasks. Furthermore, GCRL does not decrease generative performances of the decoder. Our main contributions can be summarized as follows:
\fi

\begin{itemize}
    \item We propose GCRL, a novel hybrid generative-contrastive representation learning framework. To the best of our knowledge, this is the first work to combine the generative and contrastive objectives for unsupervised representation learning on computer vision tasks.
    %{\color{blue}
    \item 
    %GCRL does not require introducing advanced architectures or inductive biases. Instead, 
    GCRL does not introduce any specialized modules or inductive biases. 
    Instead, we reinterpret the standard transformer blocks as encoder-decoder structures,  
    to which the contrastive and generative losses are separately applied, allowing us to retain the benefits of both objectives in representation learning.
    %and demonstrate that they can be trained to enjoy the benefits of both generative learning and contrastive learning without tradeoffs.}
    %\item The contrastive and generative losses are respectively applied to the transformer-based encoder and decoder, which allows to retain the benefits of both objectives in representation learning.
    \item We demonstrate that GCRL outperforms baselines on several downstream image classification tasks and out-of-distribution detection tasks and provide extensive ablation studies.
    
    %the outperformances of the proposed GCRL on several downstream image classification tasks and out-of-distribution detection with extensive ablation studies.
\end{itemize}

\section{Related Work}
\label{sec:related}

\paragraph{Self-supervised learning} Lots of self-supervised representation learning algorithms have been recently proposed for leveraging large-scale unlabeled data. In natural language processing, BERT \cite{Devlin2018arxiv} is a representative work that has exploited the masked word prediction and the next-sentence prediction as pretext tasks with bidirectional transformers and has achieved large performance improvements on many downstream tasks. Since BERT has been introduced, various variants \cite{Lan2019arxiv, Liu2019arxiv, Joshi2019arxiv} have been suggested with different pretext tasks such as the masked phrase prediction and the sentence order prediction. 
Different from BERT, InfoWord \cite{Kong2019arxiv} has been proposed to maximize the mutual information between a global sentence representation and n-grams in it, which can be considered as contrastive learning, while ELECTRA \cite{Clark2020iclr} has explored the replaced token prediction with an adversarial training. 
In computer vision, contrastive learning \cite{Oord2018arxiv} based on the manually designed positive and negative pairs has been typically used for self-supervised representation learning. % PIRL \cite{Misra2020cvpr} has tried to learn invariant features under semantic-preserving transformations. 
MoCo \cite{He2020cvpr, Chen2020arxiv} has utilized the dynamic queuing and the moving-averaged encoder for efficiently handling a large number of negative samples. SimCLR \cite{ChenT2020icml} has improved the quality of representation by finding a more proper composition of transformations and non-linear projection heads. More recently, SwAV \cite{Caron2020neurips} has modified previous pairwise representation comparisons by introducing cluster assignment and swapped prediction, and BYOL \cite{Grill2020neurips} has removed the necessity of negative pairs, bootstrapping its own representation by keeping up with the moving averaged version of itself. Here, we use SimCLR as our baseline contrastive learning algorithm due to its simplicity and powerful performance, however our method can be combined with other algorithms.

\paragraph{Generative pre-training} Since the most basic self-supervised task is to reconstruct an input itself and to maximize its likelihood, early pre-training methods were based on generative modeling. Currently, autoregressive transformers have shown state-of-the-art performances in maximizing the data likelihood, and generative pre-training (GPT) of representations using autoregressive transformers has been increasingly investigated in both natural language processing and computer vision \cite{Radford2018, Radford2019, Brown2020arxiv, Lewis2019arxiv, Chen2020icml}. Also, as generative adversarial networks (GANs) \cite{Goodfellow2014nips} have been widely used in generating high-quality images, the use of GAN for unsupervised image representation learning has been recently explored \cite{Donahue2019nips}. These generative pre-training methods generally have shown great generalization ability when combined with large-scale data and big models. The proposed GCRL build upon the image GPT (iGPT) in explicitly estimating an image likelihood.

\paragraph{Hybrid modeling} Hybrid generative and discriminative models have been largely studied to take advantages of both modeling directions \cite{Raina2004nips, Lasserre2006cvpr, Xue2010, Ricky2019nips, Mackowiak2020arxiv}, and among them, energy based models (EBMs) \cite{Grathwohl2020iclr, Liu2020arxiv, Kwonjoon2018cvpr, Yilun2019neurips} have shown great performances with improved training algorithms, recently. However, these hybrid approaches have been mostly developed for regularizing models under supervised learning, whereas our GCRL attempts to improve unsupervised representation learning. Furthermore, we separate target representations for each objective to overcome the fundamental trade-off problem in hybrid modeling.

\section{Approach}
\label{sec:approach}

This section presents how to implement GCRL without having 
trade-offs between generative and contrastive objectives.

\subsection{Hybrid Objective \& Network Architecture}

\begin{figure}[t]
  \centering
  \includegraphics[width=1.0\textwidth]{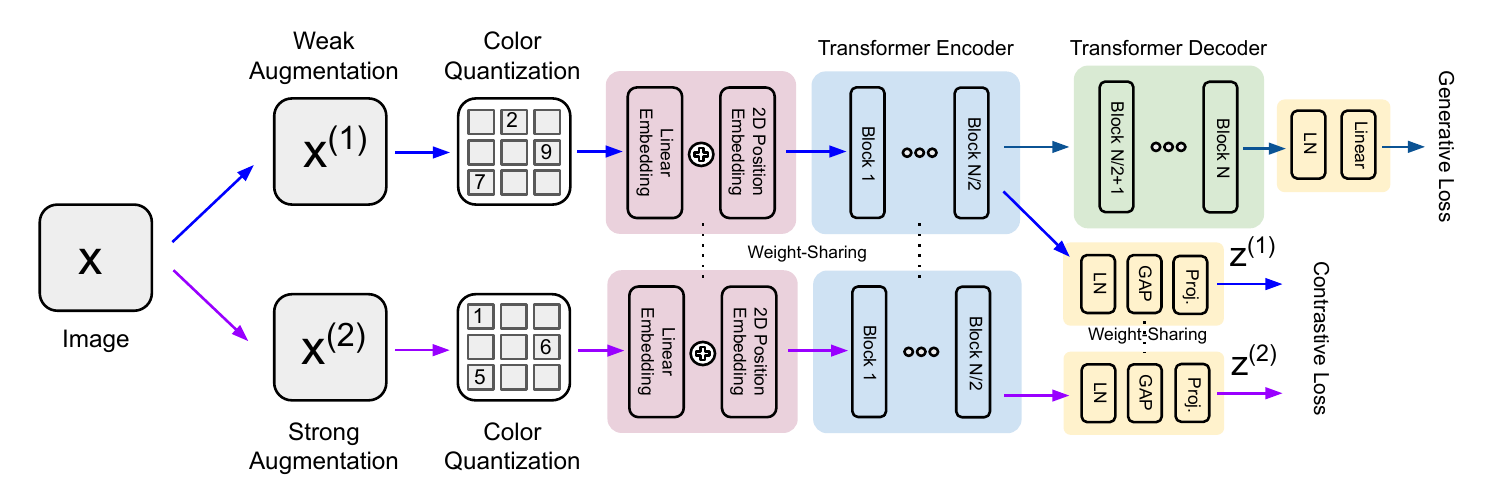}
  \caption{Illustration on GCRL architecture. 
  Here, LN and GAP stand for layer normalization and global averaging pooling. Proj. layer before the contrastive loss refers to a projection head, implemented by a two-layers neural network. We use the representation obtained before 
  the projection head for downstream classification tasks.}
  \label{fig:main_arch}
\end{figure}

Given a batch of $N$ unlabeled samples $\{\bx_i\}_{i=1}^{N}$ and the corresponding set of augmentation pairs $\{(\bx_i^{(1)}, \bx_i^{(2)})\}_{i=1}^{N}$ where $\bx_i^{(1)}$ and $\bx_i^{(2)}$ are two random augmentations of $\bx_i$, our hybrid loss function is defined as
\begin{equation}
%\label{eq:obj}
\begin{split}
    {\mathcal L} = \alpha {\mathcal L}_{g} + \beta {\mathcal L}_{c}, 
\end{split}
\normalsize
\end{equation}
where $\alpha$ and $\beta$ are the loss weights, and the generative loss ${\mathcal L}_{g}$ and the contrastive loss ${\mathcal L}_{c}$ are respectively formulated as
\small
\begin{flalign}
%\label{eq:obj}
   &{\mathcal L}_{g} \!=- \frac{1}{N}\sum_{i=1}^{N} \log p(\bx_i^{(1)}) = -\frac{1}{N}\sum_{i=1}^{N}\!\Big(\!\sum_{k=2}^{D}\log p\left(\bx_{i,k}^{(1)} | \bx_{i, <k}^{(1)} \right)\!\! \Big),\\
    &{\mathcal L}_{c} \!= - \frac{1}{2N}\sum_{i=1}^{N} \!\left(\!
        \log \!\frac{ \exp(\frac{\bz_i^{(1)} \cdot \bz_i^{(2)}}{\tau})}{\sum\limits_{j\neq i} \exp ( \frac{\bz_i^{(1)}\cdot \bz_j^{(1)}}{\tau} ) \!+\!
        \sum\limits_{j} \exp (\frac{\bz_i^{(1)}\cdot \bz_j^{(2)}}{\tau} )}
        + \log \!\frac{\exp(\frac{\bz_i^{(2)} \cdot \bz_i^{(1)}}{\tau})}{\sum\limits_{j\neq i} \exp (\frac{\bz_i^{(2)}\cdot \bz_j^{(2)}}{\tau} ) \!+\!
        \sum\limits_{j} \exp (\frac{\bz_i^{(2)}\cdot \bz_j^{(1)}}{\tau} )}
    \right).
\end{flalign}
\normalsize

% Given two streams of an input $\bx$ with different augmentation policies,
% denoted as $\bx^{(1)}$ and $\bx^{(2)}$, 
% our loss function \footnote{$N$ is the number of samples in a batch, $D$ is the data dimension, and $\tau$ means a temperature.} is described as below:
% \begin{equation*}
% %\label{eq:obj}
% \small
% \begin{split}
%     & \frac{1}{N}\sum_{i=1}^{N}\Big(\sum_{k=2}^{D}-\log p\left(\bx_{i,k}^{(1)} | \bx_{i, <k}^{(1)} \right) \Big) \\
%     & - \frac{1}{2N}\sum_{i=1}^{2N}\Big(
%         \bz_i^{(1)\top} \bz_i^{(2)} /\tau +
%         \log \big(\sum_{i,j=1}^{N} \exp (\bz_i^{(1)\top} \bz_i^{(2)}/\tau )\big)
%     \Big),
% \end{split}
% \normalsize
% \end{equation*}
Here, $D$ is the data dimension, $\bz_i^{(1)}$ and $\bz_i^{(2)}$ are projected and normalized representations extracted from $\bx_i^{(1)}$ and $\bx_i^{(2)}$, respectively, and $\bx_{i,k}^{(1)}$ is the $k$th element of $\bx_{i}^{(1)}$.
The generative loss $\mathcal{L}_g$ is the likelihood of the model autoregressively factorized in a typical raster scan order~\cite{Oord2016icml}, and the contrastive loss $\mathcal{L}_c$ is the symmetric normalized temperature
cross-entropy loss as in SimCLR~\cite{ChenT2020icml}. 

In order to optimize the loss function, we propose a transformer-based architecture composed of encoder and decoder, motivated by empirical 
observations presented in iGPT~\cite{Chen2020icml}, 
where early transformer blocks behave similarly to an encoder in typical auto-encoders and the remaining blocks behave like a decoder. 
Therefore, without altering the spatial resolution, we explicitly split a set of transformer blocks into
an encoder and a decoder as shown in Figure~\ref{fig:main_arch}. For the contrastive learning, the encoder takes two versions of an input, one is weakly augmented and the other is strongly augmented. Then, the two representations obtained from the encoder are used
for computing the contrastive loss $\mathcal{L}_c$. The decoder processes only the weakly augmented image to compute the generative loss $\mathcal{L}_g$. We observe that training of the generative loss with strong augmentations
deteriorates the robustness, since it focuses on producing
unrealistic samples from a severely distorted distribution.

\begin{table}[t]
\footnotesize
    \centering
    \caption{Model specification of each method (left). Illustration on our implementation of transformer and axial attention blocks (right). Our attention block is composed of three types of attentions: unmasked row-wise, masked column-wise, and masked row-wise attention, 
    where the parameters across attention types 
    are shared in the same block.}
    %A na\"{\i}ve implementation of this attention block requires three times more parameters than a dense attention, so we share parameters across attention types in the same block.}
    \begin{tabular}{l|ccc}
    \hline\hline
    Method & \#params & \#blocks & \#enc. blocks \\
    \hline\hline 
    \multirow{2}{*}{iGPT} & 10M & 12 & 12 \\
                          & 76M & 24 & 24 \\
    \hline 
    \multirow{3}{*}{SimCLR} & 5M & 6 & 6 \\
                            & 10M & 12 & 12 \\
                            & 76M & 24 & 24 \\
    \hline
    \multirow{2}{*}{GCRL} & (5+5)M & 12 & 6 \\
                          & (38+38)M & 24 & 12 \\
    \hline\hline
    \end{tabular}
    \hspace{.1in}
    \raisebox{-.5\height}{\includegraphics[width=0.45\textwidth]{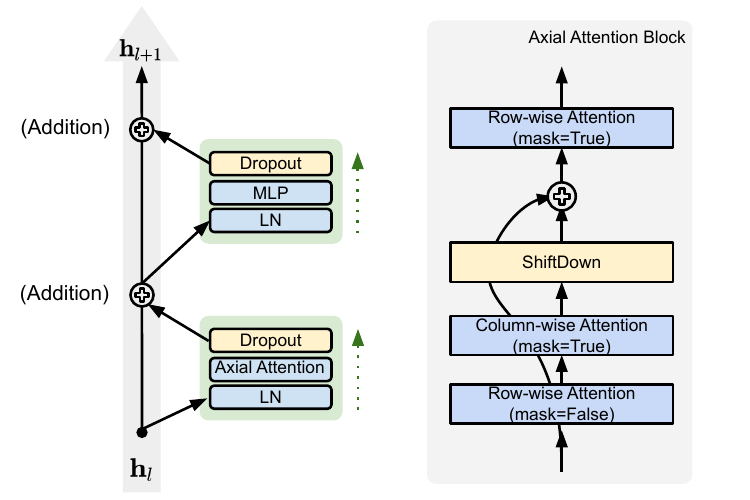}}
    \label{tab:model_spec}
\normalsize
\end{table}

\subsection{Model Details}

We use the similar transformer block as iGPT for the encoder and decoder, where each block is constructed as follows.  Let $\bh_{l-1}$ be the output from the $(l-1)$th block. 
The $l$th transformer block is composed of the following operations,
\begin{equation}
\begin{split}
    \bh_l &= \textrm{LN}(\bh_{l-1}), \\
    \bh_l &= \bh_l + \textrm{CausalMultiHeadSelfAttention}(\bh_l), \\
    \bh_l &= \bh_l + \textrm{MLP}(\textrm{LN}(\bh_l)),
\end{split}
\end{equation}
where $\textrm{LN}$ refers to layer normalization~\cite{Ba2016neurips}, causal multi-head self-attention module computes interactions across a sequence only in a raster scan order, and MLP block 
of the same structure as iGPT computes nonlinear embeddings of the tokens in a sequence independently. 
%It is noted that after unsupervised representation learning, we remove the projection head and apply the pooled representation from the encoder to downstream tasks.
To accelerate a self-attention module, we reduce the size of the sequence by color quantization~\cite{Chen2020icml} and adopt
the sparse attention block used in AxialTransformer~\cite{Ho2019arxiv} with slight modification. For color quantization, we compress RGB color channels into one of 512 codes as in iGPT, 
where the codebook is simply constructed by $k$-means clustering. 
We use the codebook provided from the official implementation of iGPT for fair comparison. %\footnote{\url{https://github.com/openai/image-gpt}}. 
Given a colored image of 32x32 resolution, this simple processing effectively reduces the sequence length from 3,072 to 1,024. 

AxialTransformer~\cite{Ho2019arxiv} accelerates a self-attention module by using a two-dimensional structure of an image, where row-wise and 
column-wise attentions are combined to access a full image in a raster scan order. 
In GCRL, unlike the original implementation of AxialTransformer, 
we sequentially apply row-wise attention, causal column-wise attention, and causal row-wise attention, to access a full image in every transformer block, as shown in Table~\ref{tab:model_spec}. We empirically observe that our implementation of axial attention block reduces working memory and training time significantly compared to  dense attention, without loosing performance. For detailed description including empirical results on the type of attention blocks, please refer to our PyTorch-like pseudo-code and ablation study in Appendix.

To fairly compare GCRL to iGPT and SimCLR, we use the same architecture for all objectives.
We perform experiments with several models by varying the number of transformer blocks (6, 12, and 24 blocks). For all models, two-dimensional positional embeddings are applied after the first linear embedding, and the number of heads for axial attention was set to 8. Table~\ref{tab:model_spec} compares the sizes of the models tested for iGPT, SimCLR, and GCRL. Note that since GCRL is composed of an encoder and a decoder, we only need a half of 
the total number of blocks for evaluating the representations. 
% so the number of encoder blocks for it in evaluating the representations is different from the total number of blocks used during the representation learning.

\section{Experiments}
\label{sec:experiments}

This section compares the characteristics of representation learned by iGPT, SimCLR, and GCRL from the perspective of linear evaluation, generative performance, robustness to out-of-distribution samples, and low-shot transfer learning.

\subsection{Experiment Details}

\begin{table}[t!]
\small
    \centering
    \caption{Pipeline of data augmentations used for SimCLR and GCRL. For GCRL, this pipeline is only applied to the input stream of heavy augmentation.}
    %\vspace{.1in}
    \begin{tabular}{c|lll}
    \hline\hline
    Order & Augmentation & Arguments & Pytorch function \\
    \hline
    1 & Resized and crop & Crop size = (32, 32), crop scale = (0.2, 1.0) & RandomResizedCrop \\
    2 & Color dist. (jitter) of prob. 0.8 & brightness=0.4, contrast=0.4, saturation=0.4, hue=0.1 & ColorJitter \\
    3 & Color dist. (gray) & probability=0.2 & RandomGrayscale \\
    4 & Horizontal flip & probability=0.5 & RandomHorizontalFlip \\
    \hline\hline
    \end{tabular} 
    \label{tab:pipeline_aug}
\normalsize
\end{table}

\begin{table}[t!]
\small
    \centering
    \caption{Detailed configurations of experiments in Table~\ref{tab:linear_eval}.}
    \begin{tabular}{cc|ccc|ccc}
    \hline\hline
    \multirow{2}{*}{Method} & \multirow{2}{*}{Model size} 
                            & \multicolumn{3}{c|}{ImageNet32} & \multicolumn{3}{c}{CIFAR10 \& 100}\\
                            & & Batch size & Epoch & Peak learning rate & Batch size & Epoch & Peak learning rate \\
    \hline\hline
    \multirow{2}{*}{iGPT}   & 10M    & 1024 & \multirow{2}{*}{100} & \multirow{2}{*}{4.0e-04} & 512 & 200 & 1.6e-03 \\ 
                            & 76M    & 384  & & & 384 & 15  & 3.2e-03 \\
    \hline
    \multirow{3}{*}{SimCLR} & 5M     & 1024 & \multirow{3}{*}{100} & \multirow{3}{*}{4.0e-04} & 512 & 400 & 4.8e-04 \\
                            & 10M    & 1024 & & & 512 & 400 & 4.8e-04 \\
                            & 76M    & 768 & & & 384 & 15  & 9.6e-04 \\
    \hline
    \multirow{2}{*}{GCRL}   & (5+5)M   & 1024 & \multirow{2}{*}{50 + 50} & \multirow{2}{*}{4.0e-04 / 4.0e-04} & 512 & 200 + 200 & 1.6e-03 / 4.8e-04  \\
                            & (38+38)M & 768 & & & 384 & 15 & 9.6e-04 \\
    \hline\hline
    \end{tabular} 
    \label{tab:hparams}
\normalsize
\end{table}

We use three datasets: 
CIFAR10, CIFAR100~\cite{cifar}, and ImageNet32~\cite{imagenet_cvpr09}, 
where CIFAR contains 50K samples for training and ImageNet32 provides 1.2M training samples of resolution 32x32.
We use Adam optimizer with decoupled weight decay (AdamW)~\cite{Loshchilov2019iclr} with 
gradient clipping of norm 1.0 and weight decay factor $10^{-4}$, and do not decay trainable parameters in layer normalization and token embeddings. 
We apply dropout with rate 0.1 before every residual connection in transformer blocks. 
We linearly increase the learning rate from zero to a specified value over 5 epochs and apply the cosine decay after it~\cite{Loshchilov2017iclr}. We use PyTorch version 1.6.0 and conduct all experiments with V100s with 32GB memories. For CIFAR experiments, we report mean and standard deviations from repetitions with random seed 0, 1, and 2. For ImageNet experiments, we report the result with random seed 0. 

We observe that it is practically useful for training GCRL only with the generative loss $\calL_g$ for the first half of the total training epochs and then train with the full objective for the remaining half. Also, we set both of the loss weights $\alpha$ and $\beta$ for the hybrid objective to $1.0$. Below, we summarize the data augmentation policies we used. 

{\bf iGPT}.
We use the standard random crop with padding followed by horizontal flip with probability 0.5 for all experiments, where the pad size is $4$ and the padding is done with reflections of images.

{\bf SimCLR}. We use the similar augmentation policy as in \cite{ChenT2020icml}. 
We use the same two-layer MLP projection head as in SimCLR, where batch normalization is replaced by layer normalization, the hidden dimension is set to 1,024, and the final embedding dimension is set to 64 for CIFAR and 128 for ImageNet experiments.

{\bf GCRL}. This requires both strong and weak augmentations where the weakly augmented images are processed for the generative loss. We choose the weak augmentation as the one used for iGPT and the strong augmentation as the one used for SimCLR. Table~\ref{tab:pipeline_aug} shows a pipeline of data augmentations for SimCLR and GCRL. We think that this augmentation policy is fair enough to see the different  behaviors of SimCLR and our approach, though it is possible to obtain more discriminative representations by employing a sophisticated augmentation strategy. 
For the projection head, we use the same network as in SimCLR. 

It is not clear what representations should be used for downstream tasks in iGPT. 
Hence, we test with two versions of representations for iGPT. The first one is pooled from the middle transformer block, and the second one is pooled from the last transformer block. For SimCLR and GCRL, the output before the projection head is selected as the final representation.

\subsection{Discriminative \& Generative Performance}
\label{subsec:dis_gen}

\begin{table*}[t]
\small
    \centering
    \caption{Comparison between baselines and GCRL in terms of linear evaluation and bits-per-dimension (bpd) on the color-quantized space. ``Pos.'' refers to which transformer block is used for the final representation of linear evaluation.}
    \vspace{.1in}
    \begin{tabular}{lcc|cc|cc|cc}
    \hline\hline
    \multirow{2}{*}{Method} & \multirow{2}{*}{Model size} & \multirow{2}{*}{Pos.} & \multicolumn{2}{c|}{ImageNet32} & 
    \multicolumn{2}{c|}{CIFAR10} & \multicolumn{2}{c}{CIFAR100} \\
    & &  &
    Acc.($\uparrow$) &
    Bpd.($\downarrow$) & 
    Acc.($\uparrow$) &
    Bpd.($\downarrow$) &
    Acc.($\uparrow$) &
    Bpd.($\downarrow$) \\
    \hline\hline
    \multirow{4}{*}{iGPT} & \multirow{2}{*}{10M} & half &
    0.2508 & \multirow{2}{*}{3.1117} &
    0.7750 {\tiny $\pm$ 0.0063} & \multirow{2}{*}{2.7775 {\tiny $\pm$ 0.0021}} & 
    0.4914 {\tiny $\pm$ 0.0054}	& \multirow{2}{*}{2.6756 {\tiny $\pm$ 0.0019}} \\
    & & last &
    0.1941 & &
    0.7482 {\tiny $\pm$ 0.0023}	&  & 
    0.4516 {\tiny $\pm$ 0.0030}	&  \\
           & \multirow{2}{*}{76M} & half &
    0.4162 & \multirow{2}{*}{3.0419} &
    0.9372 {\tiny $\pm$ 0.0024}	& \multirow{2}{*}{2.6969 {\tiny $\pm$ 0.0003}} & 
    0.7163 {\tiny $\pm$ 0.0031}	& \multirow{2}{*}{2.5978 {\tiny $\pm$ 0.0003}} \\
    & & last &
    0.3177 & & 
    0.8910 {\tiny $\pm$ 0.0014}	&  & 
    0.6171 {\tiny $\pm$ 0.0011} &  \\
    iGPT-S~\cite{Chen2020icml} & 76M & best & 0.4190 & {\bf 3.0145} & - & - & - & - \\ 
    \hline
    \multirow{3}{*}{SimCLR} & 5M & last & 
    0.2717 & \multirow{3}{*}{-} &
    0.8355 {\tiny $\pm$ 0.0018} & \multirow{3}{*}{-} & 
    0.5853 {\tiny $\pm$ 0.0058}	& \multirow{3}{*}{-} \\
           & 10M & last &
    0.2955 &  &
    0.8442 {\tiny $\pm$ 0.0064} & 	& 
    0.6022 {\tiny $\pm$ 0.0041}	&  \\
           & 76M & last &
    0.3856 &  & 
    0.9046 {\tiny $\pm$ 0.0026}	&  & 
    0.6954 {\tiny $\pm$ 0.0055} &  \\
    \hline
    \multirow{2}{*}{GCRL} & (5+5)M & half &
    0.3010 & 3.1140 &
    0.8391 {\tiny $\pm$ 0.0020}	& 2.7839 {\tiny $\pm$ 0.0023} & 
    0.5868 {\tiny $\pm$ 0.0022} & 2.6834 {\tiny $\pm$ 0.0019} \\
         & (38+38)M & half &
    {\bf 0.4359} & 3.0448 &
    {\bf 0.9506} {\tiny $\pm$ 0.0004}	& {\bf 2.6784} {\tiny $\pm$ 0.0001} &	
    {\bf 0.7603} {\tiny $\pm$ 0.0005}	& {\bf 2.5756} {\tiny $\pm$ 0.0001} \\    
    \hline\hline
    \end{tabular} 
    \label{tab:linear_eval}
\normalsize
\end{table*}

\begin{figure*}[t]
  \centering
  \begin{subfigure}
    \centering\includegraphics[width=.35\linewidth]{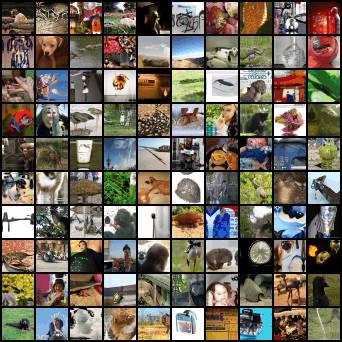}
  \end{subfigure}
  \hspace{.5in}
  \begin{subfigure}
    \centering\includegraphics[width=.35\linewidth]{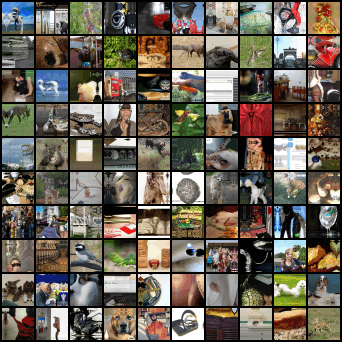}
  \end{subfigure}
  \caption{Generated samples from iGPT-76M (left) and GCRL-76M (right) trained on ImageNet32 (temperature is 1.0).
   More samples could be found in Appendix.}
  \label{fig:generated_samples}
\end{figure*}

We use the linear evaluation protocol on representations to measure the discriminative performance of visual features, and report bits-per-dimension (bpd) on 
the color-quantized space~\footnote{ Note that the bpd values reported in this paper are not directly comparable to the ones in literature~\cite{Child2019icml,Jun2020icml}, because we quantize the color space to reduce the sequence length.}
for assessing the generative performance. 
For linear evaluation, we follow the setup in the previous work~\cite{Chen2020icml,ChenT2020icml}, 
where we train a single-layer linear classifier with AdamW optimizer for 100 epochs with weight decay $10^{-4}$ and batch size 512. We use the learning rate of $0.01$ for CIFAR experiments, and $0.1$ for ImageNet experiments. No data augmentation is applied for linear evaluation.

Table~\ref{tab:linear_eval} compares GCRL with baselines under various model capacity 
and datasets, where the hyper-parameters used for this experiment are reported in Table~\ref{tab:hparams}. 
For iGPT-76M, SimCLR-76M, and GCRL-(38+38)M with CIFAR10 and CIFAR100, we fine-tune our pre-trained ImageNet32 models 
by 15 epochs. 
Table~\ref{tab:linear_eval} shows that GCRL is able to learn representations having both discriminative and generative features. When the model capacity is high enough, we observe that GCRL consistently improves the linear evaluation performance, compared to iGPT (our implementation), iGPT-S~\cite{Chen2020icml}, and SimCLR, while maintaining the generative performance. Specifically, in case of ImageNet32, 
GCRL achieves 43.6\% linear evaluation accuracy, which is superior to iGPT-S (41.9\%). 
In addition, when fine-tuning the model of 76M and (38+38)M trained on ImageNet to CIFARs, 
GCRL achieves lower bpd than the one from iGPT (2.6784 vs. 2.6969 in CIFAR10), 
which indicates the generative performance of GCRL is superior to iGPT. 

%Both classification accuracy with linear evaluation and bpd of GCRL are comparable to the ones for SimCLR and iGPT, supporting our claim that our hybrid learning objective can achieve both without tradeoffs.  Note that ours uses only the encoder (5M) for the classification, but still shows similar performance to SimCLR using twice as many parameters (10M). 

% representations in which discriminative and generative features are encoded, 
% because the performance of linear evaluation and bpd are very similar to the best of each objective. In addition, though our network requires both encoder and decoder in a training phase, we only need an encoder for classification tasks.
% Given the same inference budget, our approach performs better than SimCLR, while the generative performance is marginally dropped.

Figure~\ref{fig:generated_samples} compares the perceptual quality of generated images from 
iGPT and GCRL on ImageNet32, where both approaches are able to generate realistic images. 
In addition, we measure FID-10K~\cite{Heusel2017neurips} for iGPT and GCRL of 10M parameters trained on CIFAR10, where the score of iGPT is 45.02 $\pm$ 0.99 and the score of GCRL is 44.43 $\pm$ 0.36.
Considering these observations with the small difference between bpds in Table~\ref{tab:linear_eval}, we can argue that GCRL maintains the generative performance of iGPT.

\subsection{Robustness on OOD samples}
\label{subsec:ood}

\begin{table*}[t]
\small
    \centering
    \caption{AUROC results of a supervised OOD detection task from iGPT, SimCLR, and GCRL, 
    where we use ImageNet32, CIFAR10, and CIFAR100 as in-distribution datasets, 
    while SVHN and STL-10 are OOD datasets.}
    \vspace{.1in}
    \begin{tabular}{lcc|cc|cc|cc}
    \hline\hline
    \multirow{2}{*}{Method} & \multirow{2}{*}{Model size} & \multirow{2}{*}{Pos.} & \multicolumn{2}{c|}{ImagNet32} & 
    \multicolumn{2}{c|}{CIFAR10} & \multicolumn{2}{c}{CIFAR100}
    \\
     & & &
    SVHN &
    STL-10 &
    SVHN &
    STL-10 &
    SVHN &
    STL-10 \\
    \hline\hline
    \multirow{4}{*}{iGPT} & \multirow{2}{*}{10M} & half &
    0.9824 & 0.5458 &
    0.9638 {\scriptsize $\pm$ 0.0040} & 0.6191 {\scriptsize $\pm$ 0.0065} &  
    0.9395 {\scriptsize $\pm$ 0.0069} & 0.6214 {\scriptsize $\pm$ 0.0049} \\
           &  & last &
    0.9538 & 0.5214 &
    0.9414 {\scriptsize $\pm$ 0.0048} & 0.6311 {\scriptsize $\pm$ 0.0014} & 
    0.9236 {\scriptsize $\pm$ 0.0028} &	0.6501 {\scriptsize $\pm$ 0.0010} \\
                        & \multirow{2}{*}{76M} & half &
    0.9803 & {\bf 0.6187} &
    0.9896 {\scriptsize $\pm$ 0.0009} & {\bf 0.7325} {\scriptsize $\pm$ 0.0021} &	
    0.9697 {\scriptsize $\pm$ 0.0009} & {\bf 0.7342} {\scriptsize $\pm$ 0.0050} \\
           &  & last &
    0.9231 & 0.5243 &
    0.9261 {\scriptsize $\pm$ 0.0027} & 0.6593 {\scriptsize $\pm$ 0.0006} &	
    0.8723 {\scriptsize $\pm$ 0.0031} & 0.6797 {\scriptsize $\pm$ 0.0025} \\
    \hline
    \multirow{3}{*}{SimCLR} & 5M & last & 
    0.5867 & 0.4705 &
    0.7945 {\scriptsize $\pm$ 0.0949} & 0.4946 {\scriptsize $\pm$ 0.0076} &	
    0.8243 {\scriptsize $\pm$ 0.0728} & 0.5820 {\scriptsize $\pm$ 0.0118} \\
           & 10M & last &
    0.6415 & 0.5142 &
    0.7262 {\scriptsize $\pm$ 0.0144} & 0.4850 {\scriptsize $\pm$ 0.0055} &	
    0.7326 {\scriptsize $\pm$ 0.0355} & 0.5660 {\scriptsize $\pm$ 0.0344} \\
          & 76M & last &
    0.6196 & 0.5148 &
    0.7533 {\scriptsize $\pm$ 0.0267} & 0.5026 {\scriptsize $\pm$ 0.0107} &	
    0.7102 {\scriptsize $\pm$ 0.0163} & 0.6172 {\scriptsize $\pm$ 0.0066} \\
    \hline
    \multirow{2}{*}{GCRL} & (5+5)M & half &
    0.9975 & 0.5951 &
    0.9971 {\scriptsize $\pm$ 0.0003} & 0.6323 {\scriptsize $\pm$ 0.0052} & 
    {\bf 0.9896} {\scriptsize $\pm$ 0.0016} & 0.6375 {\scriptsize $\pm$ 0.0025} \\
                          & (38+38)M & half &
    {\bf 0.9982} & 0.6138 &
    {\bf 0.9975} {\scriptsize $\pm$ 0.0001} & 0.7000 {\scriptsize $\pm$ 0.0031} & 
    0.9841 {\scriptsize $\pm$ 0.0005} & 0.7092 {\scriptsize $\pm$ 0.0023} \\
    \hline\hline
    \end{tabular} 
    \label{tab:ood_test}
\normalsize
\end{table*}

\begin{table}[t]
\small
    \centering
    \caption{AUROC and AUPRC results of an unsupervised OOD detection task by approximating mass of a density function, where we use CIFAR10 (C10), CIFAR100 (C100), and ImageNet32 (I32) as in-distribution datasets.}
    \vspace{.1in}
    \begin{tabular}{lc|cc|cc|cc}
    \hline\hline
    \multirow{2}{*}{Method} & \multirow{2}{*}{\#params} & \multicolumn{2}{c|}{I32 $\rightarrow$ SVHN} 
                      & \multicolumn{2}{c|}{C10 $\rightarrow$ SVHN} 
                      & \multicolumn{2}{c}{C100 $\rightarrow$ SVHN} \\
           &          & AUROC & AUPRC & AUROC & AUPRC & AUROC & AUPRC \\
    \hline\hline
    \multirow{2}{*}{iGPT} & 10M & 0.9740 & 0.9847 &
                                0.8301 {\tiny $\pm$ 0.0253}	& 0.6129 {\tiny $\pm$ 0.0474} & 
                                0.8571 {\tiny $\pm$ 0.0121}	& 0.6762 {\tiny $\pm$ 0.0249} \\
                          & 76M & 0.7472 & 0.8543 &
                                0.3932 {\tiny $\pm$ 0.0093}	& 0.2222 {\tiny $\pm$ 0.0030} & 
                                0.4545 {\tiny $\pm$ 0.0062}	& 0.2502 {\tiny $\pm$ 0.0021} \\
    \hline
    \multirow{2}{*}{GCRL} & (5+5)M & 0.9710 & 0.9826 & 
                                   0.8814 {\tiny $\pm$ 0.0227}	& 0.7080 {\tiny $\pm$ 0.0484} & 
                                   0.8769 {\tiny $\pm$ 0.0094}	& 0.7167 {\tiny $\pm$ 0.0260} \\
                          & (38+38)M & {\bf 0.9950} & {\bf 0.9970} & 
                                   {\bf 0.9713} {\tiny $\pm$ 0.0022} & {\bf 0.9097} {\tiny $\pm$ 0.0064} & 
                                   {\bf 0.9565} {\tiny $\pm$ 0.0034} & {\bf 0.8754} {\tiny $\pm$ 0.0080} \\
    \hline\hline
    \end{tabular} 
    \label{tab:ood_input_grad}
\normalsize
\end{table}

We expect that our hybrid scheme learns more robust features for out-of-distribution (OOD) samples than both generative and discriminative learning since it can combine the merits of the two methods. Specifically, the generative loss is helpful for OOD discrimination because it learns the data distribution directly, and contrastive loss also helps OOD discrimination to some extent because it pulls the representations of similar in-distribution samples together to form a cluster-like structure. To validate our hypothesis, we conduct two types of OOD detection tasks using the representations learned from each objective. First, we consider a supervised OOD detection setting with the OOD detection scores that are computed from class-conditional Gaussian densities~\cite{LeeK2018neurips,winkens2020contrastive}:
%\begin{equation}
    $s(\bx)_{\textrm{sup}} = \max_{c} \left\{\calN\left({\tilde \bz}(\bx) | \boldsymbol{\mu}_c, \boldsymbol{\Sigma}_c\right) \right\}_{c=1}^{|C|}$,
%\end{equation}
where ${\tilde \bz}(\bx)$ is the representation from $\bx$ without projection, $|C|$ is the number of classes, and the parameters $(\boldsymbol\mu_c, \boldsymbol\Sigma_c)$ are estimated from 
training samples of class $c$. Second, we consider an unsupervised OOD detection setting where OOD scores are computed by area of high probability region~\cite{Grathwohl2020iclr} approximated with the magnitude of the score functions,
% by approximating the area of high probability region~\cite{Grathwohl2020iclr}, 
% using the magnitude of gradient of log-likelihood with respect to an input:
%\begin{equation}
    $s(\bx)_{\textrm{unsup}} = -\left \| \frac{\partial \log p(\bx)}{\partial \bx} \right \|_2$.
%\end{equation}
Since iGPT and GCRL quantize the input space, we cannot exactly compute this gradient with respect to the input image.
\footnote{Using a stochastic quantization~\cite{JangE2017iclr} makes it possible to derive the input gradient.}.
Instead, we compute the gradient with respect the token embedding. 
Note that SimCLR is not applicable to the unsupervised OOD detection because it does not provide a density.

\begin{figure}[t]
  \centering
  \includegraphics[width=0.16\textwidth]{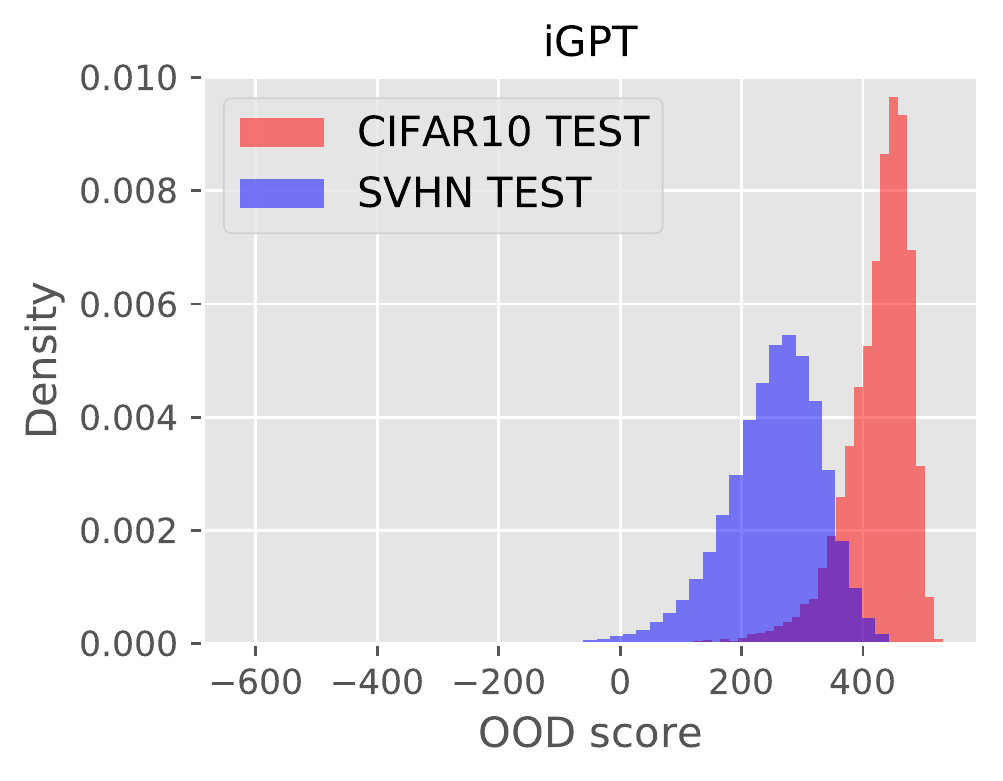}
  \includegraphics[width=0.16\textwidth]{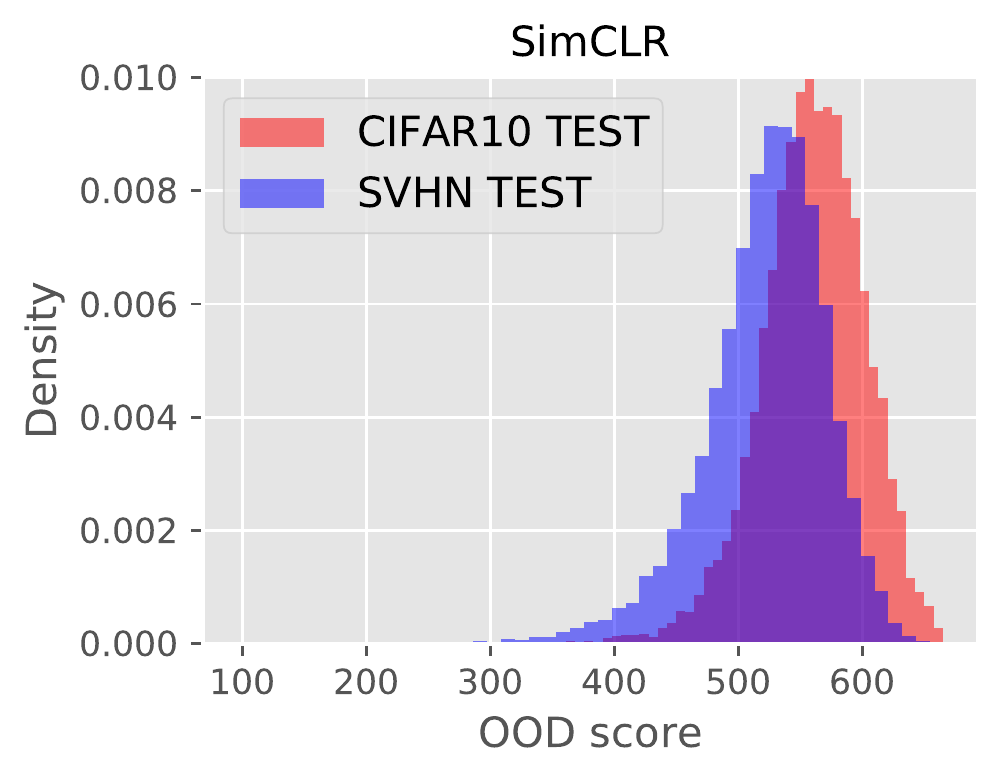}
  \includegraphics[width=0.16\textwidth]{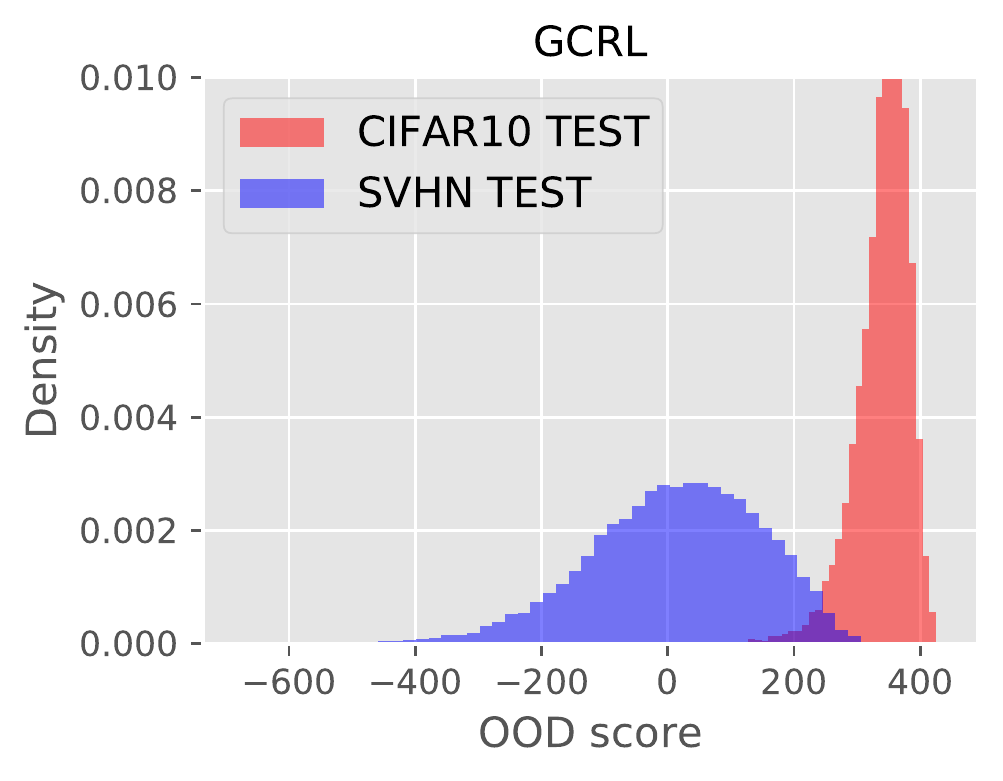}
  \includegraphics[width=0.16\textwidth]{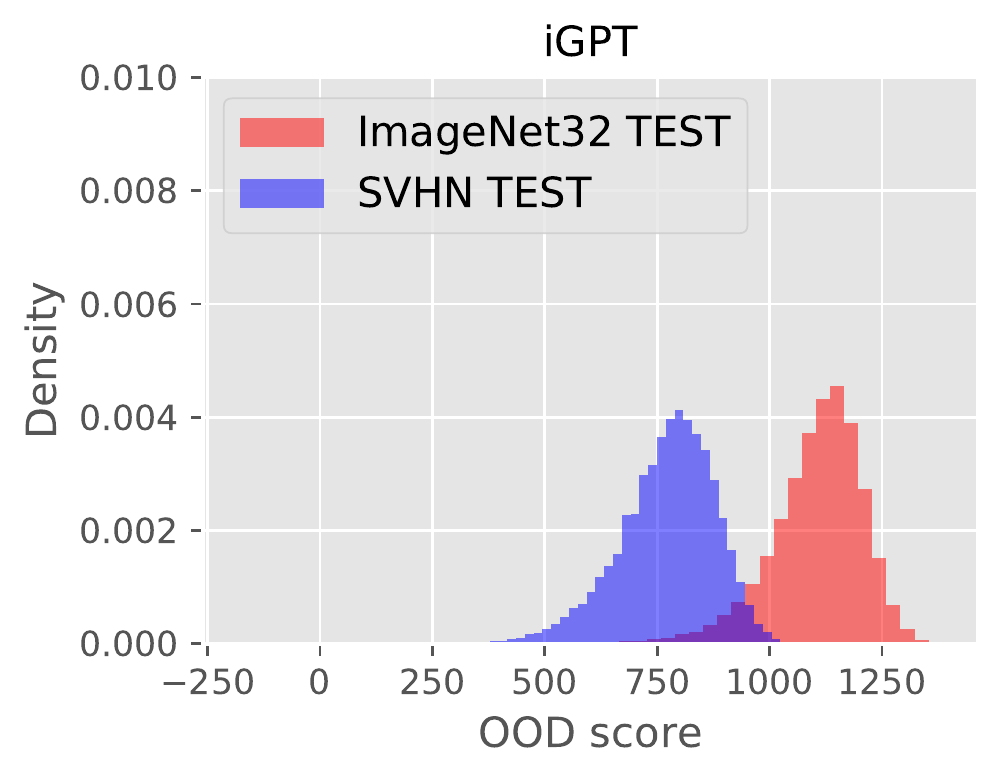}
  \includegraphics[width=0.16\textwidth]{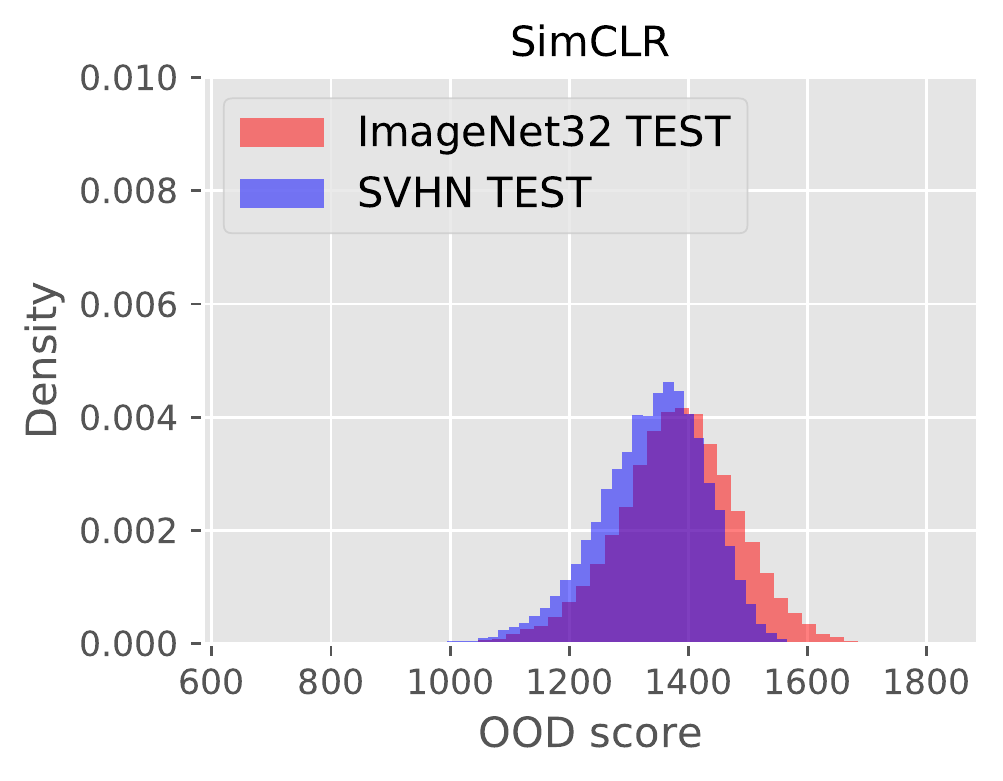}
  \includegraphics[width=0.16\textwidth]{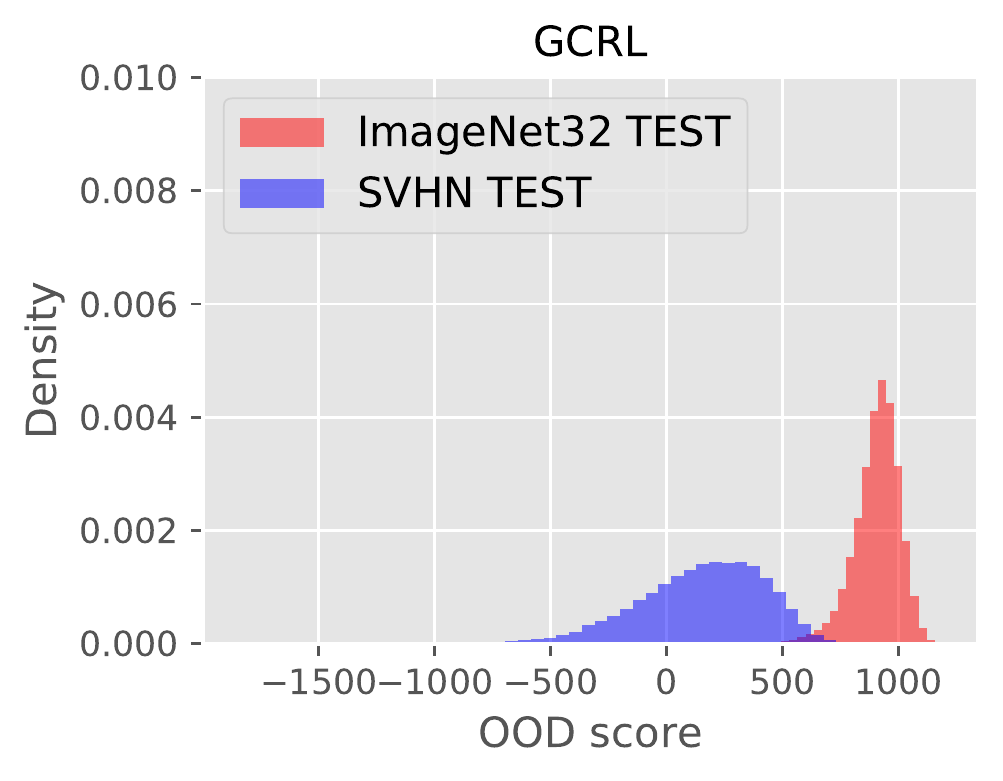}
  \caption{Distribution of supervised OOD scores of in-distribution (CIFAR10 and ImageNet32) and out-of-distribution (SVHN), where we use the 10M models for CIFAR10 and 76M models for ImageNet32. In case of iGPT, we use the representation extracted from the middle transformer block.}  \label{fig:ood_dist}
\end{figure}

\begin{figure}[t]
  \centering
  \includegraphics[width=0.22\textwidth]{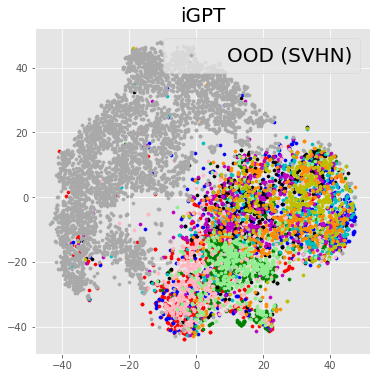}
  \hspace{.1in}
  \includegraphics[width=0.22\textwidth]{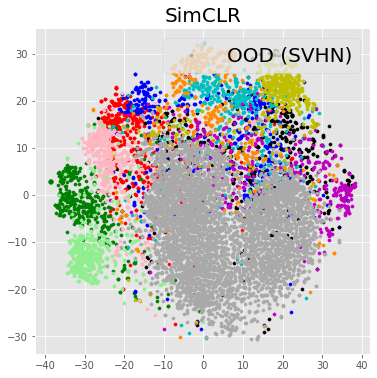}
  \hspace{.1in}
  \includegraphics[width=0.22\textwidth]{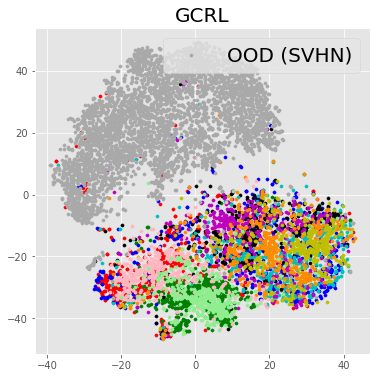}
  \caption{Two-dimensional visualization of 2,000 randomly selected test samples of CIFAR10 and SVHN, where iGPT (10M), SimCLR (10M), and GCRL of (5+5)M are trained on CIFAR10. Here, samples of SVHN are marked in gray, and samples of the same class in CIFAR10 are marked in the same color.}
  \label{fig:tsne}
\end{figure}

Table~\ref{tab:ood_test} presents the supervised OOD detection results where GCRL consistently outperforms the baselines in case of SVHN. Since STL-10 is a subset of ImageNet, it may be considered as an in-distribution set. However, we want to report how our approach behaves when a subset of in-distribution sets is given as an OOD dataset. The performance of OOD detection is measured by area under the ROC curve (AUROC), and area under the precision-recall curve (AUPRC) results are placed in Appendix. Figure~\ref{fig:ood_dist} shows several distributions of $s(\bx)_{\textrm{sup}}$ of in/out-distribution test samples, supporting the results of Table~\ref{tab:ood_test}. 
Table~\ref{tab:ood_input_grad} shows the results of the unsupervised OOD detection where GCRL outperforms iGPT. To empirically analyze the reason why GCRL works better 
than baselines, we visualize the representations learned from each method in a two-dimensional space by $t$-SNE~\cite{vanDerMaaten2008jmlr}, as shown in Figure~\ref{fig:tsne}. We can observe that iGPT is able to discriminate in-distribution and out-of-distribution samples, but the in-distribution samples are not well clustered. On the other hand, SimCLR pulls visually-similar samples together in the embedding space, but it is not able to discriminate in-distribution and out-of-distribution samples. GCRL achieves both in/out discrimination and cluster-like structure.

\subsection{Ablation Study}
\label{subsec:additional}

This section includes several additional experiments to support
our claims; (a) ablation study on the motivation of network design and low-shot transfer learning, and (b) the regularization effects of a generative loss in GCRL.

%\begin{wraptable}{r}{8cm}
\begin{table}[!t]
\small
    \centering
    \caption{Ablation study of GCRL by varying the number of blocks and the coefficients of objective on CIFAR10.}
    \begin{tabular}{ccc|cc p{1cm}}
    \hline\hline
    \#enc. / \#total & Model Size & $(\alpha, \beta)$ & Linear eval. & bpd \\
    \hline\hline
    \multirow{3}{*}{6/6} & \multirow{3}{*}{5M} & (1.0, 1.0) & 0.8034 & 2.7863\\
     &  & (0.5, 1.5) & 0.8095 &	2.8056\\
     &  & (1.5, 0.5) & 0.7929 &	2.7788\\
    \hline
    \multirow{3}{*}{12/12} & \multirow{3}{*}{10M} & (1.0, 1.0) & 0.8145 & 2.7813 \\
     &  & (0.5, 1.5) & 0.8278 & 2.7772 \\
     &  & (1.5, 0.5) & 0.8017 & 2.7872 \\
    \hline
    \multirow{3}{*}{6/12} & \multirow{3}{*}{(5+5)M} & (1.0, 1.0) & 0.8376 & 2.7866 \\
    & & (0.5, 1.5) & 0.8355 & 2.7870 \\
    & & (1.5, 0.5) & 0.8336 & 2.7887 \\ 
    \hline\hline
    \end{tabular} 
    \label{tab:abl_enc}
\normalsize
\end{table}
%\end{wraptable}

Table~\ref{tab:abl_enc} shows an ablation study to support our network design choice, showing that when the number of blocks in encoder is half of the total number of blocks, 
the final performance is robust to the hyperparameters in the GCRL objective ($\alpha$ and $\beta$).
It is noted that noticeable trade-offs between the two objectives can be observed when the hybrid loss is imposed directly to the same target representation from the final block, whereas the proposed separation of target representations can significantly reduce the trade-offs.

Table~\ref{tab:low_shot} shows the low-shot transfer classification accuracy from ImageNet32 to CIFAR10. We observe that GCRL improves the classification accuracy as well as expected calibration error (ECE)~\cite{GuoC2017icml} by a large margin. 
We believe that the generative loss greatly improves the calibration of predictions and generalization ability because it prevents overfitting to some extent, especially 
when labeled samples are scarce.

\begin{table}[t!]
\footnotesize
    \centering
    \caption{Low-shot transfer classification results of iGPT (76M), SimCLR (76M), and 
    GCRL (38+38M) with five random seeds (0,1,2,3, and 4) in case of CIFAR10.}
    \begin{tabular}{l|cc|cc|cc}
    \hline\hline
    \multirow{2}{*}{Method} &  
    \multicolumn{2}{c|}{500 labels (1\%)} & 
    \multicolumn{2}{c|}{2500 labels (5\%)} & 
    \multicolumn{2}{c}{5000 labels (10\%)} 
    \\
     & ACC. & ECE. & ACC. & ECE. & ACC. & ECE. \\
    \hline\hline
    iGPT &
           0.7417 {\tiny$\pm$0.0055} & 0.1349 {\tiny$\pm$0.0103} & 0.8332 {\tiny$\pm$0.0015} & 
           0.0566 {\tiny$\pm$0.0029} & 0.8583 {\tiny$\pm$0.0016} & 0.0353 {\tiny$\pm$0.0019} \\
    SimCLR & 
           0.7609 {\tiny$\pm$0.0029} & 0.1680 {\tiny$\pm$0.0063} & 0.8332 {\tiny$\pm$0.0013} & 
           0.0459 {\tiny$\pm$0.0023} & 0.8528 {\tiny$\pm$0.0027} & 0.0215 {\tiny$\pm$0.0033} \\
    \hline
    GCRL & 
            {\bf 0.7956} {\tiny$\pm$0.0035} & {\bf 0.0808} {\tiny$\pm$0.0086} & 
            {\bf 0.8746} {\tiny$\pm$0.0047} & {\bf 0.0260} {\tiny$\pm$0.0045} & 
            {\bf 0.8937} {\tiny$\pm$0.0027} & {\bf 0.0127} {\tiny$\pm$0.0014} \\
    \hline\hline
    \end{tabular} 
    \label{tab:low_shot}
\normalsize
\end{table}

\begin{table}[t!]
\small
    \centering
    \caption{Regularizing effects of the data likelihood in case of 
    the easy pretext tasks for a contrastive loss, where we compare GCRL of (5+5)M with SimCLR of 10M trained by a weak augmentation (WA) policy. As in Table~\ref{tab:abl_enc}, we conduct this experiment with random seed 0.}
    \raisebox{-.5\height}{\includegraphics[width=0.3\textwidth]{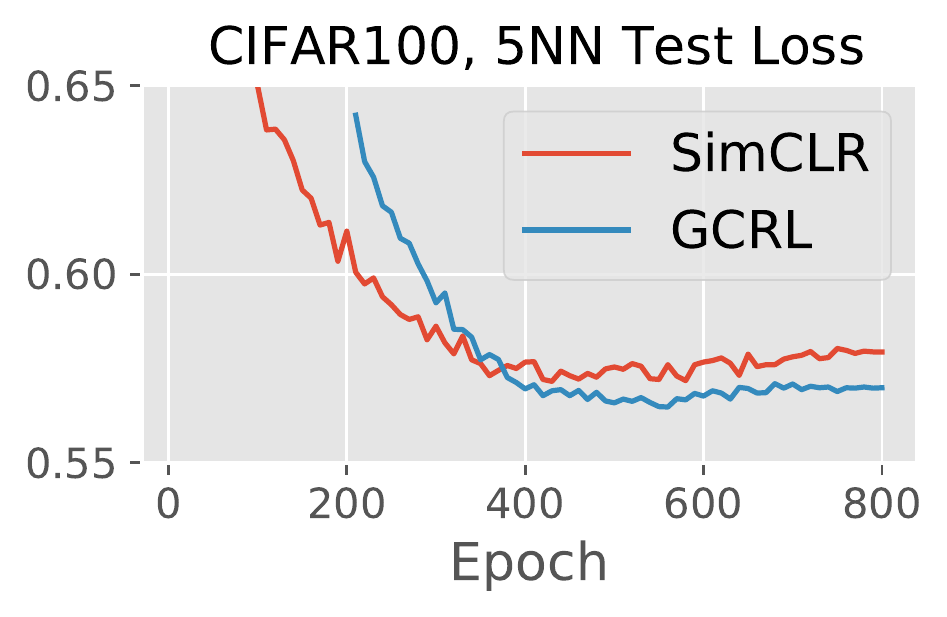}}
    \hspace{.2in}
    \begin{tabular}{lc|cc|cc}
    \hline\hline
    \multirow{2}{*}{Method} & \multirow{2}{*}{Epoch} & 
    \multicolumn{2}{c|}{CIFAR10} & 
    \multicolumn{2}{c}{CIFAR100} 
    \\
     & & ACC. & ECE. & ACC. & ECE. \\
    \hline\hline
    \multirow{2}{*}{SimCLR + WA} & 400 & 0.7414 & 0.0231 & 0.4976 & 0.0230 \\
                                 & 800 & 0.7526 & 0.0580 & 0.4763 & 0.0500\\
    \hline
    \multirow{2}{*}{GCRL + WA} & 400 & 0.8252 & 0.0145 & 0.5684 & \bf{0.0192} \\
                               & 800 & \bf{0.8331} & \bf{0.0125} & \bf{0.5789} & 0.0199 \\
    \hline\hline
    \end{tabular} 
    \label{tab:reg_cl}
\normalsize
\end{table}

\begin{table}[t!]
\small
    \centering
    \caption{Comparison of test classification accuracy and expected calibration error (ECE) between ResNet, iGPT, and GCRL on CIFAR10 and CIFAR100. 
    We train ResNet-18 by 50 epochs and iGPT by 200 epochs. We train GCRL only with a generative loss for the first 100 epochs, where the peak learning rate is 1.6e-3 and then train with the hybrid objective for the remaining 100 epochs with the peak learning rate of 9.6e-4. We repeat this experiment with random seed 0, 1, and 2.}
    \begin{tabular}{lccc|cc|cc}
    \hline\hline
    \multirow{2}{*}{Method} & \multirow{2}{*}{Size} & 
    \multirow{2}{*}{Peak LR.} &
    \multirow{2}{*}{Epoch} &
    \multicolumn{2}{c|}{CIFAR10} & 
    \multicolumn{2}{c}{CIFAR100} 
    \\
    & & & & ACC. & ECE. & ACC. & ECE. \\
    \hline\hline
    ResNet-18 & 11M & 1.0e-3 & 50 
              & 0.9360 {\tiny $\pm$ 0.0001} & 0.0445 {\tiny $\pm$ 0.0002}
              & {\bf 0.7406} {\tiny $\pm$ 0.0044} & 0.1589 {\tiny $\pm$ 0.0041} \\
    \hline
    \multirow{2}{*}{iGPT} & 5M & 4.8e-4 & 200
                          & 0.8979 {\tiny $\pm$ 0.0034} & 0.0513 {\tiny $\pm$ 0.0034} 
                          & 0.6709 {\tiny $\pm$ 0.0048} & 0.1200 {\tiny $\pm$ 0.0046} \\
                          & 10M & 4.8e-4 & 200
                          & 0.9024 {\tiny $\pm$ 0.0017} & 0.0495 {\tiny $\pm$ 0.0040} 
                          & 0.6763 {\tiny $\pm$ 0.0064} & 0.1581 {\tiny $\pm$ 0.0073} \\
    \hline
    GCRL & (5+5)M & 1.6e-3 / 9.6e-4 & 200
                  & {\bf 0.9412} {\tiny $\pm$ 0.0018} & {\bf 0.0337} {\tiny $\pm$ 0.0078} 
                  & 0.7403 {\tiny $\pm$ 0.0026}	& {\bf 0.0725} {\tiny $\pm$ 0.0104} \\
    \hline\hline
    \end{tabular} 
    \label{tab:cls}
\normalsize
\end{table}

\begin{figure}[t!]
  \centering
  \includegraphics[width=0.2\textwidth]{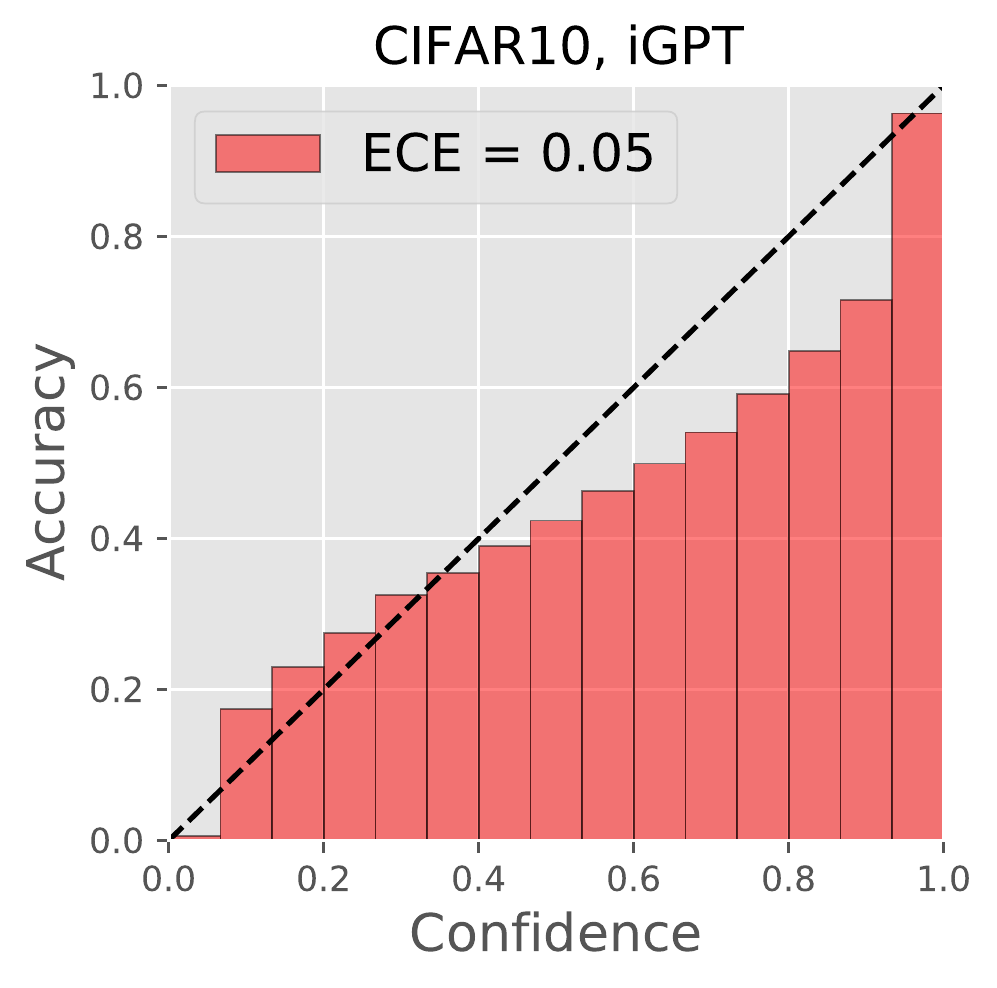}
  \hspace{.1in}
  \includegraphics[width=0.2\textwidth]{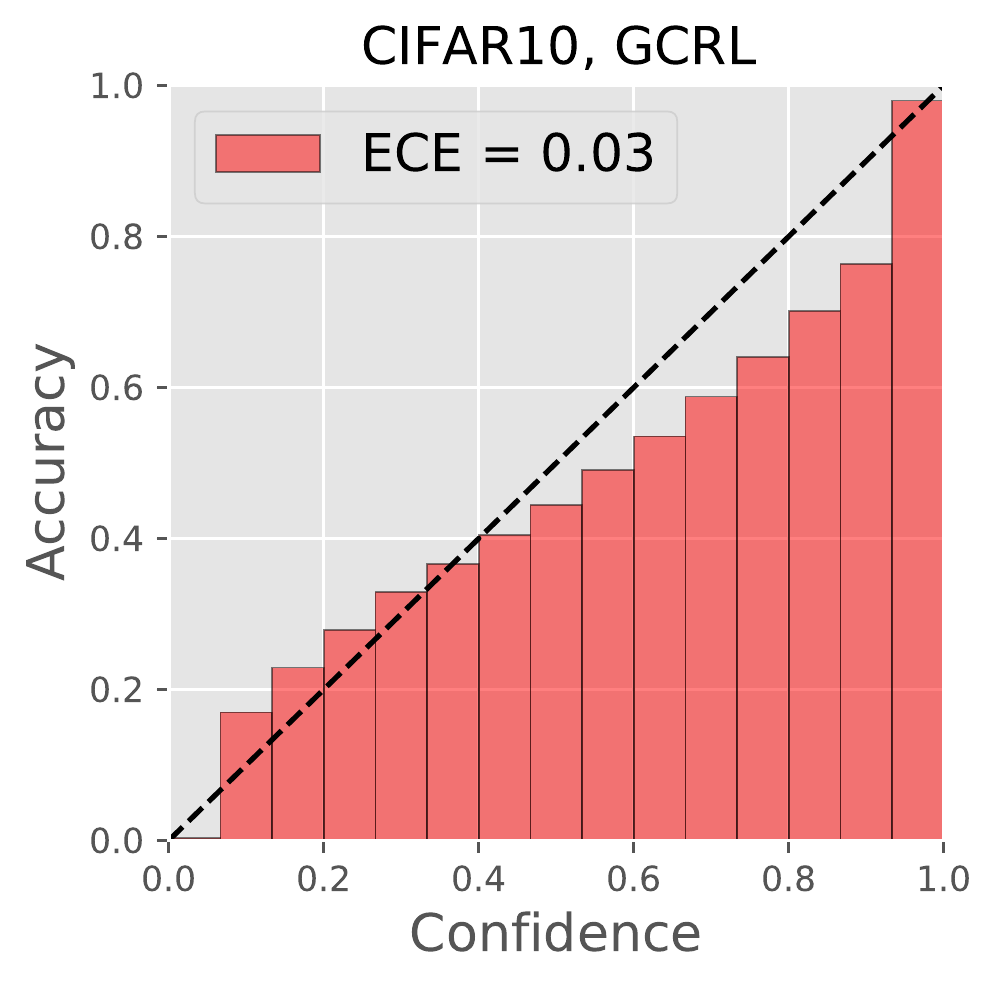}
  \hspace{.1in}
  \includegraphics[width=0.2\textwidth]{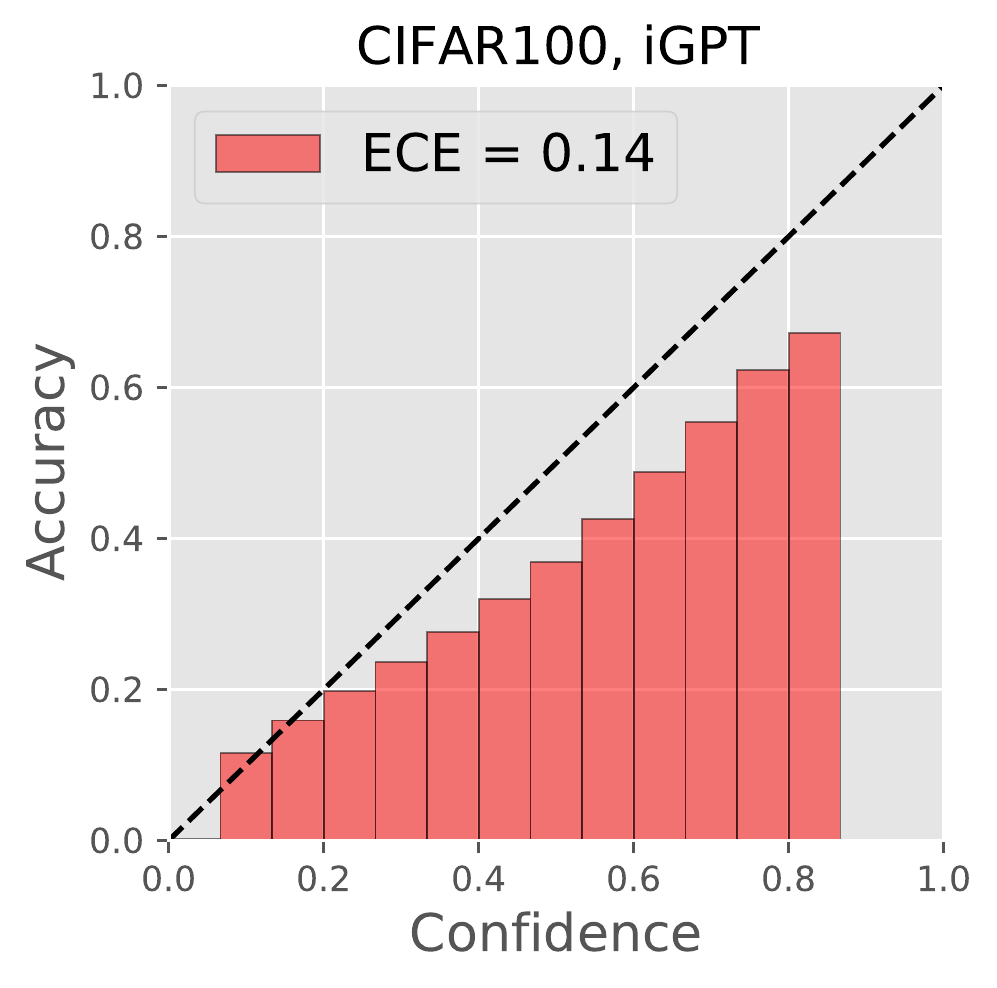}
  \hspace{.1in}
  \includegraphics[width=0.2\textwidth]{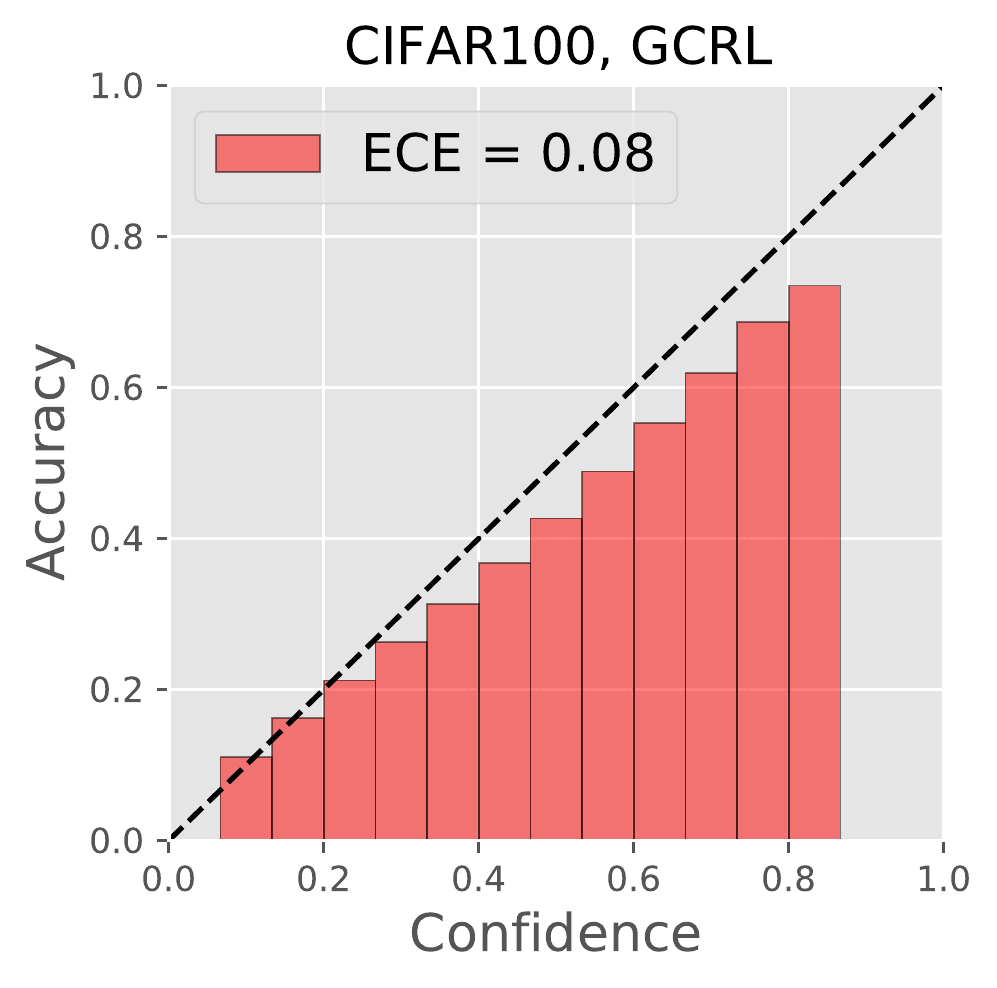}
  \caption{Reliability diagrams of iGPT and GCRL on CIFAR10 and CIFAR100, where the diagrams and ECE are  from the results of random seed 0 in Table~\ref{tab:cls}.}
  \label{fig:cls_diag}
\end{figure}

Table~\ref{tab:reg_cl} shows the regularization effect of the generative loss, where we deliberately use an easy pretext task for constrastive learning to cause overfitting. We adjust the random crop ratio from 0.2 to 0.8, meaning that at least 80\% of an original image will be included in the cropped image. 
To understand the effect of such weak augmentations for SimCLR, 
Table~\ref{tab:reg_cl} shows 5-NN classification losses of SimCLR and GCRL on test data during training. We observe that the loss of SimCLR starts increasing after 400 epochs, meaning that the model is overfitted to the pretext task, and thus hurting generalization. In addition, we observe that ECE of SimCLR at 800 epoch is higher than the one at 400 epoch, which also indicates the overfitting. 
Yet, GCRL does not suffer from such an overfitting and still performs comparable to the ones with strong augmentations thanks to the generative loss.

To show that GCRL prevents a classification model from overfitting, we conduct a supervised learning for classification with two baselines: ResNet-18~\cite{He2016cvpr} and iGPT by replacing the generative loss into the supervised cross-entropy loss.  In case of GCRL, we replace the contrastive loss into the supervised loss. 
For training ResNet, we employ a default augmentation policy~\cite{Zagoruyko2016bmvc},  where a 32x32 crop is randomly selected from a padded image with reflection of 4 pixels on each side, followed by its horizontal flip of probability 0.5. 
For GCRL as well as iGPT, we use the same augmentation policy of SimCLR in section~\ref{subsec:dis_gen}.
Table~\ref{tab:cls} shows classification accuracy and ECE of compared methods on CIFAR10 and CIFAR100. Compared to iGPT, GCRL clearly achieves the better accuracy, proving that the regularization by a generative loss is effective. 
In addition, GCRL achieves very competitive performances compared to ResNet, even though ours has minimal inductive biases in network structure.
Figure~\ref{fig:cls_diag} presents reliability diagrams of iGPT-10M and GCRL-(5+5)M models, 
showing that our approach provides well-calibrated predictions than the baseline.

\vspace{-.1in}
\section{Conclusion}
\label{sec:conclusion}
\vspace{-.05in}
We study a hybrid unsupervised training scheme that jointly optimizes generative and 
contrastive objectives in a single network. Instead of developing a complex and 
specialized module for the hybrid objective, we reinterpret the standard transformer blocks 
as an encoder-decoder structure, to which contrastive and generative losses are applied separately. We observe that our hybrid approach learns more discriminative and robust features than the ones from a single objective, especially when the model capacity is high enough. For future work, we will scale up our network to achieve more discriminative features than the ones from CNN-based architecture with recent contrastive learning approaches, while maintaining all the benefits of generative models.

\section*{Acknowledgment}
JHL was supported by Institute of Information \& communications Technology Planning \& Evaluation (IITP) grant funded by the Korea government(MSIT) (No.2019-0-00075, Artificial Intelligence Graduate School Program(KAIST))

\bibliography{refs}
\bibliographystyle{abbrv}

\newpage
\section*{Appendix}
\label{sec:appendix}

\begin{algorithm}[H]
\small
\SetAlgoLined
\# h: the representation from previous transformer block (tensor shape: BxHxWxC). Here, B refers to the batch size, (H, W) means the height and width, and C is the embedding dimension. \\
\# MSA: multi-head self-attention \\
\SetKwBlock{Begin}{def}{end def}
\Begin($\textrm{attention\_block (h):}$)
{
    h\textsubscript{a} = MSA(h.view(B*H, W, C), causal\_mask = False).view(B, H, W, C).transpose(1,2) \\
    h\textsubscript{a} = MSA(h\textsubscript{a}.view(B*W, H, C), causal\_mask = True).view(B, W, H, C).transpose(1,2) \\
    h\textsubscript{a} = torch.cat([torch.zeros(B, 1, W, C), h\textsubscript{a}[:, :-1, :, :]], axis=1) \\
    h = MSA((h + h\textsubscript{a}).view(B*H, W, C), causal\_mask = True).view(B, H, W, C) \\
    {\bf return} h
}
\caption{PyTorch-like Pseudo-code of an Axial Attention Block}
\normalsize
\label{alg:aab}
\end{algorithm}

\begin{table}[ht]
\small
    \centering
    \caption{Comparison of different attention types in iGPT (10M) on MNIST and CIFAR10 with three random seeds (0, 1, and 2).}
    \begin{tabular}{c|cc|cc}
    \hline\hline
    \multirow{2}{*}{Attention type} & \multicolumn{2}{c|}{MNIST} & \multicolumn{2}{c}{CIFAR10} \\
                                    & Throughput (images/s) & Test bpd. & Throughput (images/s) & Test bpd. \\
    \hline
    Dense                           & 303.0862 $\pm$ 0.3009 & 0.5538 {\small $\pm$ 0.0032} 
                                    & 202.2085 $\pm$ 0.2702 & 2.7676 {\small $\pm$ 0.0016} \\
    Axial                           & 383.6595 $\pm$ 0.5432 & 0.5629 {\small $\pm$ 0.0030} 
                                    & 355.0681 $\pm$ 0.1357 & 2.7775 {\small $\pm$ 0.0021} \\
    \hline\hline
    \end{tabular} 
    \label{tab:comp_axial_dense}
\normalsize
\end{table}

\begin{table}[ht]
\small
    \centering
    \caption{AUPRC results of a supervised OOD detection task, where the in-distribution datasets are ImageNet32, 
    CIFAR10 and CIFAR100.}
    \begin{tabular}{lcc|cc|cc|cc}
    \hline\hline
    \multirow{2}{*}{Method} & \multirow{2}{*}{Model size} & \multirow{2}{*}{Pos.} & \multicolumn{2}{c|}{ImageNet32}
    & \multicolumn{2}{c|}{CIFAR10} & 
    \multicolumn{2}{c}{CIFAR100} \\
     & & &
    SVHN &
    STL-10 &
    SVHN &
    STL-10 &
    SVHN &
    STL-10 \\
    \hline\hline
    \multirow{4}{*}{iGPT} & \multirow{2}{*}{10M} & half &
    0.9924 & 0.8851 &
    0.9436 {\scriptsize $\pm$ 0.0048} & 0.6715 {\scriptsize $\pm$ 0.0064} & 
    0.9158 {\scriptsize $\pm$ 0.0080} & 0.6586 {\scriptsize $\pm$ 0.0034} \\
           &  & last &
    0.9801 & 0.8775 &
    0.9208 {\scriptsize $\pm$ 0.0061} & 0.7085 {\scriptsize $\pm$ 0.0018} & 
    0.8976 {\scriptsize $\pm$ 0.0023} & 0.7079 {\scriptsize $\pm$ 0.0021} \\
                        & \multirow{2}{*}{76M} & half & 
    0.9916 & 0.9125 & 
    0.9829 {\scriptsize $\pm$ 0.0012} & 0.7931 {\scriptsize $\pm$ 0.0016} & 
    0.9491 {\scriptsize $\pm$ 0.0009} & 0.7952 {\scriptsize $\pm$ 0.0035} \\
                        & & last & 
    0.9666 & 0.8800 &
    0.9040 {\scriptsize $\pm$ 0.0031} &	0.7318 {\scriptsize $\pm$ 0.0002} & 
    0.8195 {\scriptsize $\pm$ 0.0031} & 0.7494 {\scriptsize $\pm$ 0.0029} \\
    \hline
    \multirow{3}{*}{SimCLR} & 5M & last & 
    0.7258 & 0.8522 &
    0.6211 {\scriptsize $\pm$ 0.1460} & 0.5636 {\scriptsize $\pm$ 0.0052} & 
    0.6857 {\scriptsize $\pm$ 0.1142} & 0.6754 {\scriptsize $\pm$ 0.0066} \\
           & 10M & last &
    0.7682 & 0.8691 &
    0.5276 {\scriptsize $\pm$ 0.0193} & 0.5620 {\scriptsize $\pm$ 0.0059} & 
    0.5525 {\scriptsize $\pm$ 0.0501} &	0.6687 {\scriptsize $\pm$ 0.0258} \\
           & 76M & last & 
    0.7706 & 0.8646 &
    0.6310 {\scriptsize $\pm$ 0.0337} & 0.5743 {\scriptsize $\pm$ 0.0105} & 
    0.5573 {\scriptsize $\pm$ 0.0150} & 0.7065 {\scriptsize $\pm$ 0.0064} \\
    \hline
    \multirow{2}{*}{GCRL} & (5+5)M & half &
    0.9989 & 0.9061 &
    0.9938 {\scriptsize $\pm$ 0.0006} & 0.7026 {\scriptsize $\pm$ 0.0038} & 
    0.9803 {\scriptsize $\pm$ 0.0027} & 0.7018 {\scriptsize $\pm$ 0.0025} \\
         & (38+38)M & half & 
    0.9992 & 0.9090 &
    0.9953 {\scriptsize $\pm$ 0.0002} & 0.7600 {\scriptsize $\pm$ 0.0030} & 
    0.9698 {\scriptsize $\pm$ 0.0011} & 0.7820 {\scriptsize $\pm$ 0.0015} \\
    \hline\hline
    \end{tabular} 
    \label{tab:sup_ood_auprc}
\normalsize
\end{table}

We introduce a PyTorch-like pseudo-code for our implementation of axial attention blocks and compare the performance of it to the dense attention blocks. 
Algorithm~\ref{alg:aab} shows how to transform the representation from previous block in the axial attention block. 
We remark that all parameters across attention types in the same block are shared.

Table~\ref{tab:comp_axial_dense} compares our axial attention with dense attention 
in iGPT of 10M parameters in terms of inference throughput and test bpd. 
We train the models on MNIST and CIFAR10 by 30 and 200 epochs, respectively. 
We do not apply data augmentations to MNIST. For CIFAR10, we apply horizontal flipping of probability 0.5 and random cropping with resizing of pad size 4. We measure inference throughput by a V100 GPU with 100 repetitions, where we use the batch sizes of 48 and 24 for MNIST and CIFAR10 in case of the dense attention. 
We use the batch sizes of 128 and 64 for MNIST and CIFAR10 in case of the axial attention block.
Table~\ref{tab:comp_axial_dense} shows that our axial attention block performs comparably to the dense attention block, 
while improving inference throughput by 26\% (MNIST) and 75\% (CIFAR10). 
We remark that it is free to choose other sparse attention implementations, since the main contribution of our work is to introduce the benefits of representation learned from a hybrid objective in an unsupervised fashion.

\begin{figure}[t]
  \centering
  \includegraphics[width=0.8\textwidth]{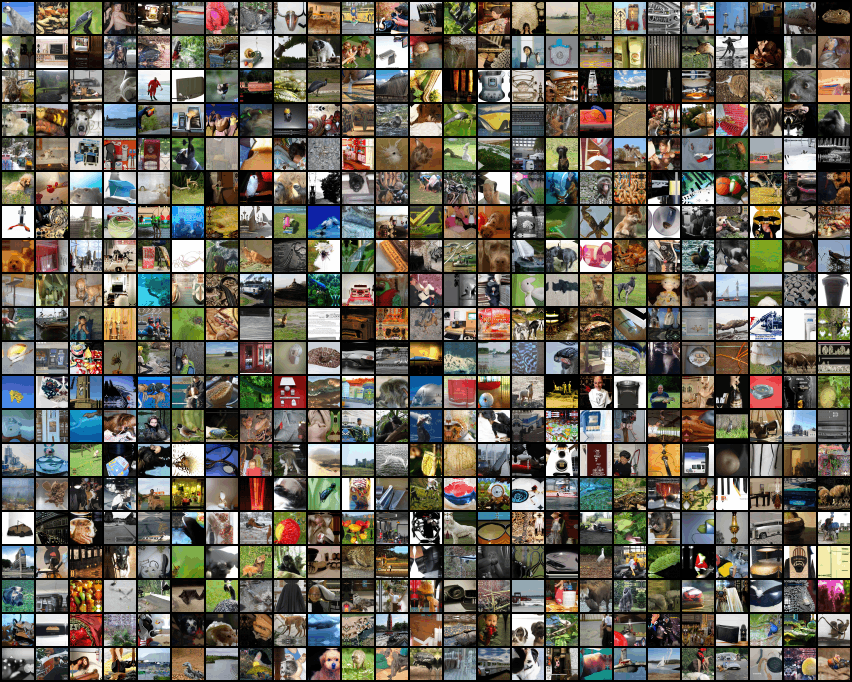}
  \caption{Generated images from GCRL of (38+38)M trained on ImageNet32 ($\tau = 1.0$).}
  \label{fig:gen_temp_1.0}
\end{figure}

\begin{figure}[t]
  \centering
  \includegraphics[width=0.8\textwidth]{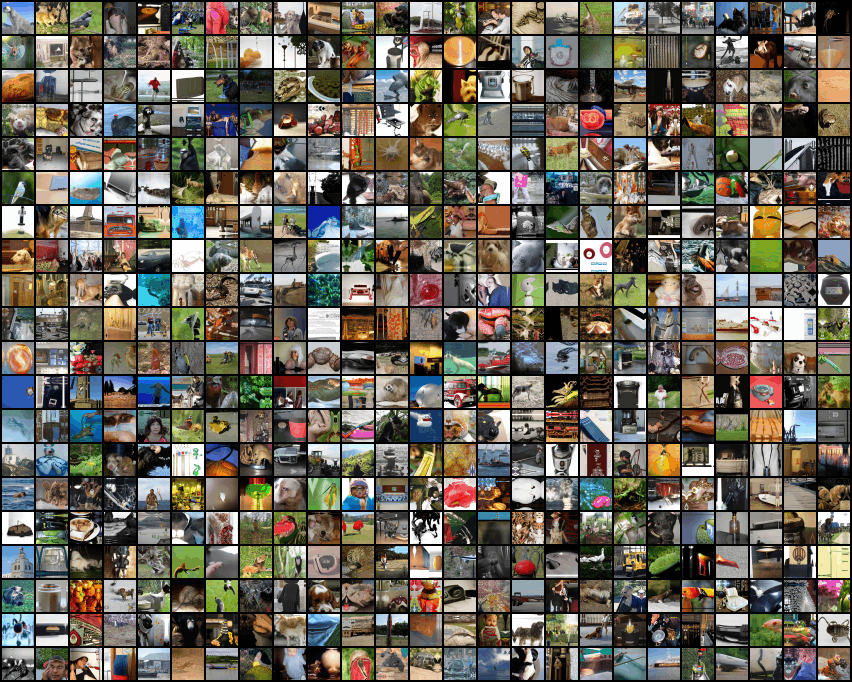}
  \caption{Generated images from GCRL of (38+38)M trained on ImageNet32 ($\tau = 0.99$).}
  \label{fig:gen_temp_0.99}
\end{figure}

\begin{figure}[t]
  \centering
  \includegraphics[width=0.8\textwidth]{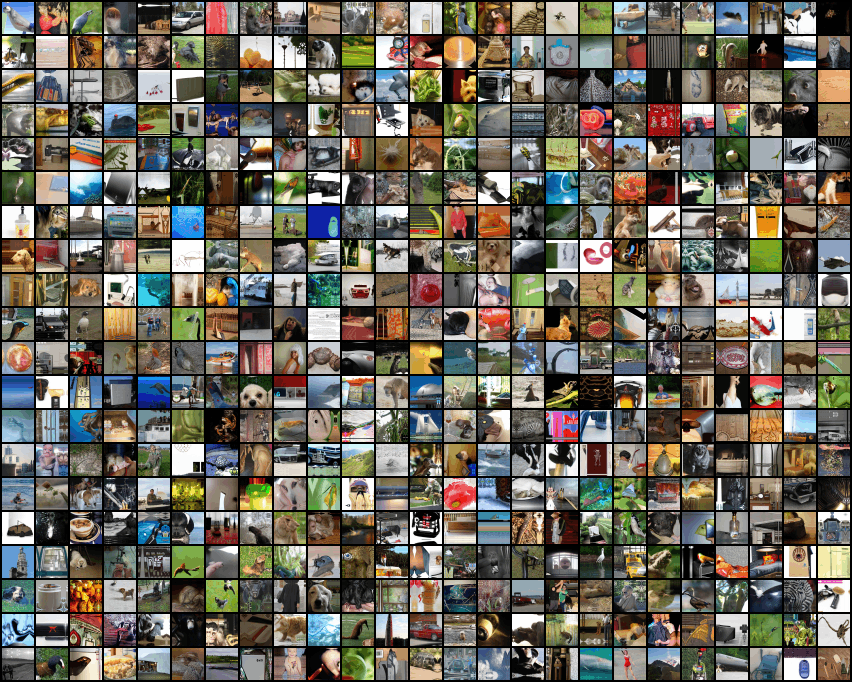}
  \caption{Generated images from GCRL of (38+38)M trained on ImageNet32 ($\tau = 0.98$).}
  \label{fig:gen_temp_0.98}
\end{figure}

Table~\ref{tab:sup_ood_auprc} presents AUPRC results of the supervised OOD detection task, 
showing the same trend of AUROC results.
Figure~\ref{fig:gen_temp_1.0}-\ref{fig:gen_temp_0.98} show 500 generated samples with softmax temperature (denoted as $\tau$) of 1.0, 0.99, and 0.98, where the same random seed is used.
Regardless of the temperature, our model is able to generate images of high perceptual quality. 

%\bibliography{refs}
%\bibliographystyle{icml2021}

\end{document}